\documentclass[10pt,twocolumn,letterpaper]{article}

\usepackage[pagenumbers]{iccv} 

\definecolor{iccvblue}{rgb}{0.21,0.49,0.74}
\usepackage[pagebackref,breaklinks,colorlinks,allcolors=iccvblue]{hyperref}
\usepackage[accsupp]{axessibility}  

\def\paperID{7190} 
\def\confName{ICCV}
\def\confYear{2025}

\title{OpenM3D: Open Vocabulary Multi-view Indoor 3D Object Detection
\\without Human Annotations}

\usepackage{orcidlink}
\author{
  Peng-Hao Hsu$^{1}$\thanks{Equal contribution.}, 
  Ke Zhang$^{2}$\footnotemark[1], 
  Fu-En Wang$^{2}$, 
  Tao Tu$^{3}$,
  Ming-Feng Li$^{4}$, 
  Yu-Lun Liu$^{5}$, \\
  Albert Y. C. Chen$^{2}$, 
  Min Sun$^{1,2}$\thanks{Equal advisory contribution.},
  Cheng-Hao Kuo$^{2}$\footnotemark[2]\\
  \textsuperscript{1}National Tsing Hua University \quad
  \textsuperscript{2}Amazon \quad
  \textsuperscript{3}Cornell University \quad \\
  \textsuperscript{4}Carnegie Mellon University \quad
  \textsuperscript{5}National Yang Ming Chiao Tung University\\
}

\usepackage{graphicx}	
\usepackage{amsmath}	
\usepackage{amssymb}	
\usepackage{booktabs}
\usepackage{microtype}
\usepackage{makecell}
\usepackage{epsfig}
\usepackage{float}
\usepackage{placeins}
\usepackage{color, colortbl}
\usepackage{stfloats}
\usepackage{enumitem}
\usepackage{tabularx}
\usepackage{xstring}
\usepackage{multirow}
\usepackage{xspace}
\usepackage{url}
\usepackage{subcaption}
\usepackage{xcolor}
\usepackage{fixmath}
\usepackage{ifthen}

\usepackage[linesnumbered,ruled]{algorithm2e}

\usepackage{fixmath}
\usepackage{booktabs}
\usepackage{enumitem}

\newcommand{\ourmethod}{OpenM3D}

\newcommand{\innersec}[1]{\vspace{3pt}\noindent\textbf{#1}}

\newcommand{\mdeq}[1]{Eq.~#1}

\newcommand{\mdvec}[1]{\mathbold{#1}}
\newcommand{\mdmat}[1]{\mathbold{#1}}

\newcommand{\mdineq}[1]{\(#1\)}

\newboolean{revising}
\setboolean{revising}{false}
\ifthenelse{\boolean{revising}}{
    \newcommand{\peng}[1]{\textcolor{violet}{[peng]: #1}}
}{
    
    \newcommand{\peng}[1]{#1}
    \newcommand{\ke}[1]{#1}
    \newcommand{\minsun}[1]{#1}

}

\usepackage{xr-hyper}

\makeatletter
\newcommand*{\addFileDependency}[1]{
  \typeout{(#1)}
  \@addtofilelist{#1}
  \IfFileExists{#1}{}{\typeout{No file #1.}}
}

\makeatother

\begin{document}
\maketitle

\begin{abstract}
Open-vocabulary (OV) 3D object detection is an emerging field, yet its exploration through image-based methods remains limited compared to 3D point cloud-based methods. We introduce {\ourmethod}, a novel open-vocabulary multi-view indoor 3D object detector trained without human annotations. In particular, {\ourmethod} is a single-stage detector adapting the 2D-induced voxel features from the ImGeoNet model. To support OV, it is jointly trained with a class-agnostic 3D localization loss requiring high-quality 3D pseudo boxes and a voxel-semantic alignment loss requiring diverse pre-trained CLIP features. We follow the training setting of OV-3DET where posed RGB-D images are given but no human annotations of 3D boxes or classes are available. We propose a 3D Pseudo Box Generation method using a graph embedding technique that combines 2D segments into coherent 3D structures. Our pseudo-boxes achieve higher precision and recall than other methods, including the method proposed in OV-3DET. We further sample diverse CLIP features from 2D segments associated with each coherent 3D structure to align with the corresponding voxel feature. The key to training a highly accurate single-stage detector requires both losses to be learned toward high-quality targets. At inference, {\ourmethod}, a highly efficient detector, requires only multi-view images for input and demonstrates superior accuracy and speed (0.3 sec. per scene) on ScanNet200 and ARKitScenes indoor benchmarks compared to existing methods. We outperform a strong two-stage method that leverages our class-agnostic detector with a ViT CLIP-based OV classifier and a baseline incorporating multi-view depth estimator on both accuracy and speed. 
\end{abstract}
\vspace{-5mm}
\section{Introduction}
\label{sec:intro}


Thanks to the recent breakthrough of Vision-Language Models (VLMs)~\cite{CLIP,ALIGN,OpenSeg}, general representations aligned across the 2D image and free-form-text spaces have become available. These VLMs demonstrate impressive generalization ability to zero-shot object classification tasks. A line of work~\cite{gu2022openvocabulary,xu2022simple,ghiasi2022scaling} combines existing class-agnostic 2D object proposals with the zero-shot ability of VLMs to classify 2D object proposals into a large number of object classes.
These methods open the door for Open-Vocabulary (OV) 2D object detection and segmentation, handling free-form text descriptions for objects at inference time.
For robotics applications, another line of work explores OV 3D indoor scene understanding~\cite{conceptfusion,ha2022semantic,Peng2023OpenScene,takmaz2023openmask3d,lu2023openvocabulary} based on lifting image features from VLMs to 3D. 
However, all these methods require high-quality 3D point cloud as inputs. 
This reliance on expensive 3D sensors (\eg, depth cameras, stereo cameras, or laser scanners) is the bottleneck.
On the other hand, for fixed-vocabulary, several multi-view image-based methods~\cite{rukhovich2022imvoxelnet,xu2023nerfdet,tu2023imgeonet} have achieved significantly improved 3D object detection performance. Unlike point cloud-based methods, image-based methods do not require expensive 3D sensors at inference time.


We propose {\ourmethod}, a novel OV multi-view indoor 3D object detector that trained without human annotations. To the best of our knowledge, this is the first work generalizing the OV capability to multi-view 3D object detection.
{\ourmethod} is a single-stage 3D detector adapting the 2D-induced voxel features from the ImGeoNet model. The voxel feature is aggregated from multiple RGB features.  To support OV, we need to enable it to localize all objects and classify them according to OV descriptions.
Hence, it is jointly trained with a class-agnostic 3D localization loss requiring high-quality 3D pseudo boxes and a voxel-semantic alignment loss requiring diverse pretrained CLIP features. We follow the training setting of OV-3DET where posed RGB-D images are given but no human annotations of 3D boxes or classes are available. We proposed a 3D Pseudo Box Generation method using a graph embedding technique that combines 2D segments into coherent 3D structures during training (See Fig.~\ref{fig:gen_pseudo_box}). Specifically, we apply SAM~\cite{SAM} on multi-view images to obtain class-agnostic 2D segments. By treating each segment as a node and computing the relation between nodes according to their connectivity in 3D, we formulate the 3D pseudo bounding boxes generation as a novel graph embedding-based clustering problem so that segments connected in 3D are clustered into a 3D object instance.
Our pseudo boxes achieve higher precision and recall than other methods, including those proposed in OV-3DET. We further sample diverse CLIP features from 2D segments associated with each coherent 3D structure to align with the corresponding voxel feature. The key to training a highly accurate single-stage detector requires both losses to be learned toward high-quality targets. While depth is utilized during pseudo-box generation and training, it is not needed for inference.
At inference time, 
{\ourmethod} is a single-stage OV 3D object detector that requires only multi-view RGB images as input and runs in 0.3 seconds per scene on a V100 GPU. In contrast, most 3D scene understanding methods necessitate point clouds or depth information, as well as the large CLIP ViT model to be applied, leading to significantly higher computational costs. For example, OV-3DET~\cite{lu2023openvocabulary} takes 5 seconds per scene, while OpenMask3D~\cite{takmaz2023openmask3d} requires 5–10 minutes per scene.


We evaluate {\ourmethod} on ScanNet200~\cite{rozenberszki2022language} and ARKitScenes~\cite{arkitscenes}. Our 3D pseudo-boxes achieve higher accuracy than those from OV-3DET~\cite{lu2023openvocabulary} and SAM3D~\cite{yang2023sam3d} by jointly considering 2D segments across all views and employing graph embedding-based clustering to mitigate frame-wise errors.
{\ourmethod} also outperforms its counterparts trained with OV-3DET’s and SAM3D’s 3D boxes in both class-agnostic and multi-class 3D object detection on ScanNet200, demonstrating the effectiveness of our 3D pseudo-boxes.
Moreover, {\ourmethod} with 3D voxel representation surpasses a strong two-stage baseline that classifies objects using 2D CLIP ViT features on both datasets, validating the contribution of Voxel-Semantic feature alignment.
We also compare against a multi-view depth estimation baseline, which first estimates depth, applies graph embedding-based clustering for 3D box proposals, and classifies objects using 2D CLIP ViT features. This approach is at least 270 times slower due to depth estimation and CLIP ViT inference, while {\ourmethod} achieves superior mAP and mAR.




The contributions of our work are the following.
\begin{itemize}[itemsep=1pt, topsep=0pt, parsep=0pt]
\item {\ourmethod} is the first multi-view open-vocabulary 3D object detector achieving SoTA accuracy on ScanNet200 and ARKitScenes. 
\item A novel Voxel-Semantic Alignment loss is proposed to align 3D voxel features with multi-view CLIP embeddings. This loss enables open-vocabulary classification by aggregating diverse CLIP features from multiple viewpoints, capturing different object appearances across angles. 
\item {\ourmethod} is then trained jointly with both localization and alignment losses as a single-stage detector running 0.3 seconds per scene on V100.
\item For localization loss supervision, we propose a novel 3D pseudo box generation pipeline that leverages graph embedding to integrate 2D segments into a coherent 3D structure, surpassing existing methods in experiments.
\end{itemize}

\section{Related Work}
\label{sec:related}

\noindent{\textbf{3D Object Detection.}}3D object detection in indoor scenes has gained more research attention due to the availability of datasets with ground truth 3D bounding boxes~\cite{arkitscenes,dai2017scannet}. When the 3D point cloud is available at inference time, two types of methods are proposed to leverage the 3D geometric information. 
Point-based methods directly sample based on set abstraction and feature propagation~\cite{qi2017pointnet++, qi2018fpointnet, yang2019std, shi2019pointrcnn, yang20203dssd, shi2020pointgnn, pan2021pointformer, qi2019votenet, zhang2020h3dnet}, while grid-based methods are based on grid representation~\cite{yang2018pixor, zhou2018voxelnet, yan2018second, lang2019pointpillars, shi2020pillarod, deng2021voxelrcnn, mao2021voxeltransformer, gwak2020gsdn, rukhovich2022fcaf3d}.
Despite the fact that point cloud-based methods perform well on object detection, they rely on costly 3D sensors, which narrows down their use cases.

\noindent{\textbf{Multi-View 3D Object Detection.}} When 3D point clouds are not available at inference time, several other methods can leverage multi-view RGB images for 3D object detection.
DETR-based approaches~\cite{wang2022detr3d, liu2022petr, tseng2022crossdtr} expand upon the capabilities of DETR~\cite{carion2020detr} to tackle the challenge of 3D object detection. Previous studies~\cite{huang2021bevdet, li2022bevformer} have established the effectiveness of the bird-eye-view (BEV) representation for object detection in autonomous driving scenarios. 
Another approach focuses on constructing 3D feature volumes from 2D observations. ImVoxelNet~\cite{rukhovich2022imvoxelnet} achieves strong indoor 3D object detection using a voxel-based feature volume~\cite{murez2020atlas}, but struggles to preserve the intrinsic scene geometry. NeRF-Det~\cite{xu2023nerfdet} addresses this by integrating NeRF to estimate 3D geometry while minimizing latency through geometry priors and a shared MLP for a geometry-aware volume. Concurrently, ImGeoNet~\cite{tu2023imgeonet} introduces a geometric-shaping component that predicts surface structure from multiple RGB images in the feature volume and enhances geometric precision. In this work, we build our open-vocabulary multi-view 3D object detector on top of the geometric-shaped 3D feature volume introduced in ImGeoNet.

\noindent\textbf{2D OV Detection.} Open-vocabulary object detection (also known as zero-shot detection) is the task of detecting novel classes for which no training labels are provided~\cite{gu2022openvocabulary, RahmanAAAI20,ZareianCVPR21}. Recent methods~\cite{gu2022openvocabulary} employ image-text pairs to extract rich semantics from text, thus expanding the number of classes of the detector. However, the detector classes will be fixed after training. Another solution is to replace the classifier with pre-trained vision-language embeddings~\cite{RahmanAAAI20,ZareianCVPR21}, allowing a detector to utilize an OV classifier and perform OV detection.

\noindent\textbf{3D OV Detection.} PointCLIP~\cite{zhang2021pointclip} accomplishes OV recognition of point clouds by projecting them into multi-view images and processing these images with CLIP~\cite{CLIP}. However, this method cannot be directly applied to point-cloud detection because it does not handle the localization of unknown objects.
Recently, OV-3DET~\cite{lu2023openvocabulary} proposed a 3D point cloud-based 3D object detector learning to align point cloud-based feature with pre-trained CLIP~\cite{CLIP} feature space. They leverage a large-scale pre-trained external OV-2D detector~\cite{Detic} to generate 3D pseudo boxes for potential novel objects. Since OV-3DET applies OV-2D detector at each view to generate 3D pseudo boxes, there are a large number of overlapping 3D boxes compared to our proposed method.
\minsun{Moreover, CoDA~\cite{cao2023coda} tackles 3D OV detection in a different setting. It assumes a set of base classes are available with ground truth 3D boxes. Then, an iterative novel object discovery and model enhancement procedure is proposed. Hence, we compare with 3D pseudo boxes from OV-3DET rather than CoDA due to differences in training settings.}
\peng{ImOV3D~\cite{yang2024imov3d} mitigates the scarcity of annotated 3D data in OV 3D object detection by generating pseudo-multimodal representations from 2D images to bridge the modality gap with 3D point clouds.}
However, all these methods rely on 3D point clouds during inference, whereas our proposed method is a multi-view image-based 3D object detector.


\noindent\textbf{3D OV Scene Understanding.}
\minsun{Beyond 3D OV detection, many works in 3D scene understanding have been recently proposed. 
OpenScene~\cite{Peng2023OpenScene} is the seminal work aligning the representation of 3D points with CLIP features in the posed images from back-projection. However, it does not support the output of object 3D box or 3D segment explicitly.
OpenMask3D~\cite{takmaz2023openmask3d} is designed for 3D instance segmentation. It projects 3D instance mask proposals to 2D posed images and refines them with SAM. Label predictions are made by comparing the CLIP features on visual masks and text prompts. 
However, both OpenScene and OpenMask3D require point clouds and CLIP model computation for inference.
LeRF~\cite{kerr2023lerf} fuses multi-scale CLIP features extracted from 2D multi-view images into a neural radiance field for OV queries. Although no point clouds are required, an extra scene reconstruction step and CLIP model computation are required at inference time. In comparison, {\ourmethod} is an efficient single-stage OV 3D object detector only requiring multi-view images as input and runs 0.3 seconds per scene during inference.}
\section{Preprocess: 3D Pseudo Box Generation}
\label{sec.m-3dbox}
\begin{figure*}[t!]
\vspace{-1.5em}
\begin{center}
  \includegraphics[width=\textwidth]{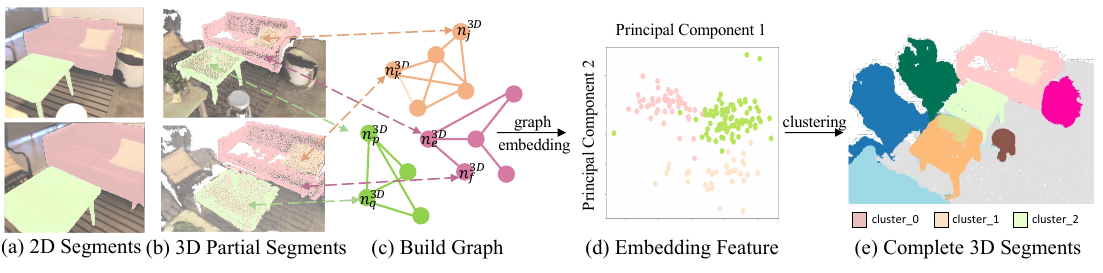}
\end{center}
\vspace{-1.7em}
\caption{
  \textbf{Graph Embedding-Based 3D Pseudo Box Generation.} Given multi-view images, we utilize an off-the-shelf class-agnostic 2D instance segmentation approach to produce 2D segments $S^{2D} = \{n^{2D}_{j}\}$ (see color-coded 2D masks in (a)). Each 2D segment is then lifted in 3D to form a partial 3D segment $n^{3D}_{j}$ following \mdeq{\ref{eq:projection}} (see (b)). Next, we build a graph in which each partial segment $n^{3D}_{j}$ is a node and we determine edges based on the overlap of segments in 3D following \mdeq{\ref{eq:bw_warp}} (see (c)). The graph embedding feature is computed for each node based on the graph (see (d)).
  Finally, nodes are clustered with the embedding features to yield complete 3D segments (see (e)). \textit{Best viewed in color.}
}
\vspace{-1em}
\label{fig:gen_pseudo_box}
\end{figure*}
Our training follows OV-3DET~\cite{lu2023openvocabulary}, eliminating manual 3D annotations by leveraging 2D vision and vision-language models to associate class information with posed RGB-D images. To train a class-agnostic detector with high recall and precision, we generate 3D pseudo-boxes from these images by (1) lifting 2D segments into partial 3D segments using a class-agnostic 2D segmenter and (2) merging them across viewpoints via a graph embedding-based method to form complete 3D segments. Fig.~\ref{fig:gen_pseudo_box} illustrates the proposed 3D pseudo-box generation process.

%

\noindent\textbf{Partial 3D Segments (Fig.~\ref{fig:gen_pseudo_box}~(a,b)).}
Given an RGB image $I$, we extract the 2D segments $n_{j}^{\text{2D}}$ of the whole scene $S^{\text{2D}}=\{n_{j}^{\text{2D}}\}$ using an off-the-shelf class-agnostic 2D instance segmentation approach, where each 2D segment contains a set of pixel positions $\{(u, v)_q\}$. 
Subsequently, we lift each 2D segment into 3D based on the provided camera pose $(\mdmat{R}, \mdmat{t})$, intrinsic $\mdmat{K}$, and depth map $\mdmat{D}$ by
\begin{equation} \label{eq:projection}
    \begin{bmatrix}
        x \\
        y \\
        z 
    \end{bmatrix}
    =\mdmat{R}^T \mdmat{K}^{-1} \mdmat{D}(u,v)
    \begin{bmatrix}
        u \\
        v \\
        1 
    \end{bmatrix}
    -
    \mdmat{R}^T \mdmat{t}~.
\end{equation}
We denote the result of \mdeq{\ref{eq:projection}} as a partial 3D segment $n^{\text{3D}}_j = \{(x, y, z)_q\}$, comprising numerous 3D points $(x, y, z)$, as it only includes partial observation from a single view. 
To incorporate information from multiple viewpoints, we explore methods for aggregating 3D partial segments, 
though this process is complicated by noise from imperfect 2D segments. 
The goal is to merge these noisy partial segments into a more robust 3D representation. 
While aggregating partial segments across views is intuitive for completing a 3D object's surface, 
achieving this without amplifying noise and bias remains challenging.

\noindent\textbf{Complete 3D Segments (Fig.~\ref{fig:gen_pseudo_box}~(c,d,e)).}
A simple sequential aggregation of partial segments accumulates errors due to incomplete object understanding. It relies on limited consecutive frames from a previous time step, resulting in inherent noise.
Hence, we propose a graph embedding-based method that considers all viewpoints simultaneously.
The scene is represented as a graph, with each node as a partial 3D segment $n^{3D}_j$, and edges indicating a high likelihood on nodes of the same object.

%
%
To learn such graph representation, we apply an off-the-shelf graph embedding method on the graph data that considers the entire scene.
%
%
For a pair of nodes (\ie, partial segments) in this graph data, an edge is established when the overlapping ratio between two nodes exceeds a specific threshold \mdineq{\theta} as follows:
%
\begin{align} \label{eq:bw_warp}
    &e_{jk} = \mathrm{edge}(n_{j}^{\text{3D}}, n_{k}^{\text{3D}}) = \begin{cases}
        1, & \text{if \mdineq{O(n_{j}^\text{3D}, n_{k}^\text{3D})}} > \theta,\\
        0, & \text{else}
    \end{cases} ~, \\
\label{eq:overlap}
&O(n_{j}^\text{3D}, n_{k}^\text{3D}) = \frac{|n_{j}^{\text{3D}}\bigcap n_{k}^{\text{3D}}|}{\text{min}(|n_{j}^{\text{3D}}|, |n_{k}^{\text{3D}}|)}
\end{align}
%
where $|\cdot|$ counts the number of points in a set, and $\bigcap$ denotes intersection of two sets.
We only consider the node pairs within the same voxel for efficiency.
The final graph shows the interconnections between nodes (\ie, overlapping partial 3D segments) across the entire scene from multiple views.

%
%
After obtaining node features from the off-the-shelf graph embedding method, we generate complete 3D segments that account for partial object segments from all viewpoints by grouping similar nodes (\ie, partial segments) using K-Means.
We then form a complete 3D segment by collecting all partial segments in the same cluster $q$ as follows $\hat{n}^{\text{3D}}_{q} = \{n^{\text{3D}}\in \mathcal{C}_{q}\},$
where $\mathcal{C}$ is the set of partial 3D segments that share the same segment index of a clustered group $q$.

\innersec{Mesh Segmentation Refinement.}
Besides our complete 3D segments derived from multi-view images, we can further consider the 3D segments $S^{\text{mesh}} = \{n^{\text{mesh}}_{j}\}$ that are generated by a mesh-based segmentation method.
%
Using ground truth mesh as input, we apply an off-the-shelf graph cut method~\cite{felzenszwalb2004efficient} to generate an additional set of 3D segments $S^{\text{mesh}}$.
To fuse these two kinds of 3D segments, that is $\{n^{\text{mesh}}_{j}\}$ and $\{\hat{n}^{\text{3D}}_{q}\}$, we apply \mdeq{\ref{eq:overlap}} to determine the overlapping ratio between any pairs of two segments from images and the mesh.
%
For each 3D segment $n^{\text{mesh}}_{j}$ from the mesh, we identify its overlapped 3D segment from our complete 3D segments $\hat{n}^{\text{3D}}_{h}$ with the highest overlapping ratio, and subsequently update the segment index in $n^{\text{mesh}}_{j}$ from $j$ to $h$.
By updating segment indices and combining 3D segments with the same segment index, we fuse the over-segmentation of the mesh to refine the original complete 3D segments back-projected from multi-view.

\innersec{3D Boxes from Complete 3D Segments.} 
To derive the axis-aligned 3D bounding box that encompasses each complete 3D segment $\hat{n}^{3D}_q$, 
we calculate the box center $(x,y,z)$ as the mean 3D position of the 3D segment in each axis direction, and the minimum and maximum coordinates in each direction to derive the width $w$, length $l$, and height $h$ of the 3D box.
%
Additionally, we apply thresholding on the volume of each 3D box and the number of points contained within each box to remove abnormally small or less visible boxes.

\begin{figure*}[t]
\vspace{-1em}
\begin{center}
  \includegraphics[width=0.9\textwidth]{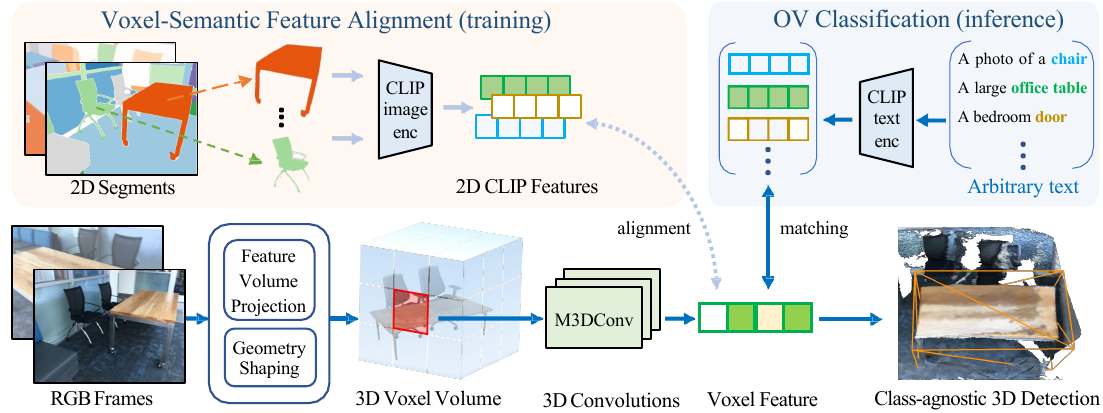}
\end{center}
\vspace{-1.5em}
\caption{
\textbf{Overview.} {\ourmethod} learns class-agnostic 3D box prediction and open-vocabulary (OV) assignments during training and only needs multi-view RGB images to infer OV 3D boxes. The bottom branch is our class-agnostic 3D object detector (Sec.~\ref{sec.m-CA3D}), where we build the 3D voxel features based on ImGeoNet. During training, given a set of RGBD images and their corresponding poses, we back-project 2D features from images to form the initial 3D voxel volume and perform geometry-shaping on the volume for 3D object localization. The top-left panel is our Voxel-\minsun{Semantic} feature alignment branch (only required in training) to empower OV classification on the 3D voxel features (Sec.~\ref{sec.m-align}). In the training phase, we use a depth map to match \minsun{semantic features from 2D segments extracted by CLIP encoder to corresponding 3D voxels and align the semantic features} and voxel features. In the inference phase, {\ourmethod} only requires a set of RGB images and their corresponding camera poses to predict 3D boxes and perform OV classification, as shown in the top-right panel.
}
\vspace{-1em}
\label{fig:model}
\end{figure*}
\section{OpenM3D}
\label{sec:method}
\peng{{\ourmethod}, our multi-view 3D object detector, is trained using posed RGB-D images from diverse indoor scenes. It employs a class-agnostic 3D localization loss (Sec.\ref{sec.m-CA3D}) supervised by 3D pseudo boxes and a voxel-semantic feature alignment loss (Sec.\ref{sec.m-align}) to enable OV classification by aligning voxel features with pre-trained CLIP features. During inference, only RGB images and their corresponding camera poses are required.}
\subsection{Class-Agnostic 3D Localization Loss}\label{sec.m-CA3D}

Given the 3D pseudo boxes with no class labels, we train a class-agnostic multi-view 3D object detector based on the model architecture of ImGeoNet~\cite{tu2023imgeonet}.
We recap the model architecture and introduce the learning target below.

The input to the class-agnostic detector consists of a sequence of images \mdineq{\mdmat{I}_t} along with camera intrinsics \mdineq{\mdmat{K}} and extrinsics \mdineq{\mdmat{P}_t}. 
We back-project the 2D features of these images to construct a 3D feature volume \mdineq{\mdmat{V} \in \mathbb{R}^{H_v \times W_v \times D_v \times C}}, where \mdineq{H_v}, \mdineq{W_v} and \mdineq{D_v} represent the height, width, and depth of the 3D volume in terms of the voxel size unit. The channel dimension of each voxel in the 3D volume is denoted as \mdineq{C}. 
Each voxel feature is further weighted according to the probability of that voxel being located on an object's surface to incorporate geometry-shaping information. 
We define the likelihood of the voxel being on an object surface as the geometry shaping volume \mdineq{\mdmat{S}}, and the geometry shaping network \mdineq{geo(\cdot)} to generate $\mdmat{S} = geo(\mdmat{V}^{'})$. Here we concatenate feature variance with the feature volume \mdineq{\mdmat{V}} to obtain \mdineq{\mdmat{V}^{'}}. Note that \mdineq{\mdmat{S}} shares the same volume size as the original feature volume \mdineq{\mdmat{V}}. As such, the geometry-aware feature volume is obtained by directly applying the geometry shaping weights from \mdineq{\mdmat{S}} to the original feature volume ($\mdmat{V}_{geo} = \mdmat{S} \odot \mdmat{V})$~.
  Furthermore, we add dense 3D convolution layers to the geometry-aware volume to harvest volumes at different scales, which has proven helpful in detecting objects at different sizes~\cite{rukhovich2022imvoxelnet, tu2023imgeonet}: $\{\mdmat{V}_h^{(i)} = \text{Conv3D}^{(i)}(\mdmat{V}_{geo}) \ |\ i \in \{0,1,...,L-1\}\}~.$
  In total we have \mdineq{L} volumes at different scales.
A single-stage anchor-free detector is deployed as the detection head, and takes the multiscale feature volume $\mdmat{V}_h^{(i)}$ as input. We train each voxel cell to predict the 3D pseudo box location using rotated 3D IoU loss~\cite{zhou2019iou} for aligning box center, size, and yaw; centerness using cross-entropy loss~\cite{tian2020fcos}, which reflects the proximity of the voxel to object centers; and binary foreground probability using focal loss~\cite{lin2017focal} to address the foreground-background imbalance. The complete process of the class-agnostic detector is depicted as the bottom branch in Fig.~\ref{fig:model}.

\subsection{Voxel-Semantic Feature Alignment Loss}\label{sec.m-align}
The 3D feature volume \mdineq{\mdmat{V}_h} acquired through class-agnostic 3D object detection encodes rich geometric information for 3D object localization. It still lacks the capability of open-vocabulary classification on 3D objects.
To address this limitation, we introduce an OV classification branch atop the class-agnostic 3D object detection branch during training. 
We propose a novel training loss to minimize the difference between voxel features from \mdineq{\mdmat{V}_h} and pre-trained CLIP features extracted from the projected multi-view 2D segments (See Fig.~\ref{fig:model}, Top-Left Panel). We refer to this as the Voxel-Semantic Feature Alignment Loss, leveraging CLIP’s ability to capture image-text semantics and bridge the gap between 2D visual concepts and 3D voxel representations.
In detail, we first deploy the image encoder of CLIP to extract the embedding \mdineq{\mdvec{f}^{2D}_j} for each 2D segment $n^{2D}_j$. Following a similar strategy of Eq.\ref{eq:projection} in Sec.~\ref{sec.m-3dbox}, we lift each 2D segment $n^{2D}_j$ to a partial 3D segment $n^{3D}_j = \{(x, y, z)_q\}$. Each 3D point \mdineq{(x, y, z)} is then mapped to a voxel indexed by \mdineq{{H_v \times W_v \times D_v}} in \mdineq{\mdmat{V}_h}. The set of voxel indices for \mdineq{n^{3D}_j} is denoted as \mdineq{V^{3D}_j = \{(x_k, y_k, z_k)_{k}\}}, and each corresponding voxel feature is denoted as \mdineq{\mdvec{f}^{3D}_k = \mdmat{V}_h(x_k, y_k, z_k, :)\in \mathbb{R}^C}. The training objective is to minimize the cosine distance between CLIP embeddings from 2D segment and voxel features from 3D volume for alignment:
$\mathcal{L}_{align}(\mathbf{\mdvec{f}_j^{2D}},\{\mdvec{f}^{3D}_k\}) = \sum_{k \in V^{3D}_j}(1 - \frac{\mathbf{\mdvec{f}^{2D}_j} \cdot \mathbf{\mdvec{f}^{3D}_k}}{\|\mathbf{\mdvec{f}^{2D}_j}\| \|\mathbf{\mdvec{f}^{3D}_k}\|})~,$
where the $j^{th}$ 2D segment is aligned with each voxel feature indexed by $k \in V^{3D}_j$.

\noindent\textbf{Inference.}  
{\ourmethod} simply gets 3D boxes from the class-agnostic 3D detector, and computes softmax over cosine similarity (i.e., matching) between average voxel features within the box and the CLIP text embeddings of text prompts to perform OV classification (See Fig.~\ref{fig:model}-Top-Right Panel). 
Since CLIP text embeddings are precomputed, {\ourmethod} is a single-stage detector not requiring heavy computation of CLIP visual or text features during inference.
\minsun{Contrastingly, most 3D scene understanding methods necessitate the computation of the large CLIP ViT model during inference.}
\section{Experiments}
\label{sec:exp}
We first introduce the setup including datasets, evaluation metrics, and baseline methods (Sec.~\ref{sec.data_eval}), and implementation details (Sec.~\ref{sec.details}).
In Sec.~\ref{sec:exp-pseudo}, we compare 3D pseudo boxes generated by our proposed {\ourmethod} to ones by the current state-of-the-art approaches, OV-3DET~\cite{lu2023openvocabulary} and SAM3D~\cite{yang2023sam3d}. For 3D object detection, we show our experimental results in Sec.~\ref{sec:exp-ov}. We further conduct ablation studies of our proposed {\ourmethod} in Sec.~\ref{sec:exp-ablation}.
\subsection{Datasets, Evaluation Metrics, and Baselines}\label{sec.data_eval}

\noindent\textbf{ScanNet200.} 
The ScanNet dataset~\cite{dai2017scannet} is a widely used RGB-D video dataset for benchmarking various 3D tasks. There are in total 1,513 room-level scenes, with 2.5 million views. We use 1,201 scenes for training and 312 scenes for testing, which adhere to the public train-val split proposed in ScanNet.
Following~\cite{qi2019votenet, tu2023imgeonet}, we generate axis-aligned bounding boxes based on semantic labels assigned to the 3D mesh of each scene. For the training and evaluation of image-based methods, a uniform sampling of 20 views per scene is performed, guided by frame indices. The images are then standardized to a resolution of 480 \mdineq{\times} 640. 
Rozenberszki \etal~\cite{rozenberszki2022language} further extend ScanNet from 18 to 200 object categories, denoted as \textbf{ScanNet200}. The 200 categories are further split into 3 subsets, based on the frequency of the number of labeled surface
points in the train set, namely head (66 classes), common (68 classes), and tail (66 classes) groups. 
Our experiment focuses primarily on the ScanNet200 dataset due to the diversity of its categories, aligning well with real-world OV scenarios. Note that there are 189 classes available on the validation set for our evaluation.

\noindent\textbf{ARKitScenes.} ARKitScenes~\cite{arkitscenes} provides 5,048 scans collected from 1,661 scenes using Apple LiDAR sensors. These scans contain RGB-D frames along with 3D object bounding box annotations of 17 categories. Due to the limited categories compared to ones in ScanNet200, ARKitScenes is used to evaluate recall rather than precision.

Besides, it is noteworthy that for the 3D object detection task, the point clouds in ARKitScenes are of lower quality compared to ScanNet200, as the depth maps in ARKitScenes are low-resolution (192×256) from iPad Pro.

\noindent\textbf{Evaluation Metrics.} 
We employ typical precision and recall for evaluating 3D pseudo box performance on ScanNet200 training set. Moreover, the average precision (AP) and the average recall (AR) metrics are also applied to measure detection performance. For each class, AP is calculated by computing the area under the precision-recall curve, and AR is computed as the average recall across all intersection over union (IoU) thresholds. More precisely, we utilize \mdineq{AP_{25}} and \mdineq{AR_{25}}, where the numerical values denote the 3D IoU threshold as $0.25$, the minimum IoU required to classify a detection as a positive match to a ground truth box. Consequently, we report mean AP and mean AR across all classes, denoted as \mdineq{mAP_{25}} and \mdineq{mAR_{25}}.

\noindent\textbf{Baseline - Pseudo 3D Boxes.} 
We generate pseudo-3D boxes from the segmentation results of SAM3D~\cite{yang2023sam3d}, which requires frames from multiple viewpoints as input. We also use OV-3DET~\cite{lu2023openvocabulary} generated boxes independently from each frame, which results in a large number of near-duplicated pseudo boxes. Hence, it has low precision and longer training times compared to our approach. To mitigate this issue, we randomly sample the same number of pseudo boxes generated by our method.

\noindent\textbf{Baseline - Strong Two-stage Detector (S2D).} Unlike {\ourmethod}, a single-stage detector simultaneously localizing 3D objects and classifying open-set descriptions of objects, S2D uses {\ourmethod} to localize candidate 3D boxes. In the second stage, each candidate 3D box is projected back to multi-view 2D images, and the CLIP embeddings are extracted. The averaged embedding over all projected regions is used to match text prompts of each class. Note that this baseline uses CLIP ViT/B-32 during the inference stage which is 7 times slower than our method.

\noindent\textbf{Baseline - \minsun{S2D using} Depth estimated 3D Boxes.}
We employ a well-trained multi-view depth estimator \cite{Bae2022} to extract depth from testing images, and generate 3D bounding boxes \peng{by pseudo box generation using estimated depth}. The second stage of the Strong Two-Stage approach classifies these boxes, providing an effective solution for open-vocabulary multi-view 3D object detection. However, an inference time of 81 seconds per scene for depth estimation on a V100 GPU is prohibitively long, making it impractical for real-world applications.
\subsection{Implementation Details}\label{sec.details}
\noindent\textbf{Class-agnostic 2D Segments and CLIP Embeddings.} Given multi-view RGB images, we generate class-agnostic 2D instance segments by SAM~\cite{SAM}. Besides, to mitigate the impact of background elements such as the floor, walls, and ceiling, which can potentially introduce errors in graph embedding due to their substantial spatial presence, we use \cite{li2022mask} to filter out such backgrounds in training. We then extract each segment with CLIP image encoder for the embedding. 
Unless otherwise specified, we utilized ViT/L-14 in our experiments, with minimal impact observed from alternative image encoders like ViT/B-32 and ViT/B-16. For more details, please refer to our supplementary material. 

\noindent\textbf{Coordinates Standardization.}
To address minor variations in 3D point coordinates caused by depth map noise, we standardize coordinates across all 3D partial segments. We employ voxelization and K-nearest neighbors (KNN) to fuse 3D points to vertices extracted from the ground truth mesh. This procedure, involving the fusion of point sets within voxel grids to the nearest extracted vertex through KNN, ensures a unified representation of coordinates in 3D partial segments from diverse viewpoints.
%
%

\noindent\textbf{Complete 3D segments and boxes.}
We construct a graph from partial 3D segments and apply DeepWalk~\cite{perozzi2014deepwalk} to generate graph embeddings. We then cluster these embeddings using K-means (K=100 for all scenes), grouping similar nodes into clusters that form complete 3D segments. To assign a segment label to each point, we first map every node’s cluster index to its associated 3D points. For points shared across multiple nodes, we apply majority voting to determine the final segment assignment. Additionally, a connected-component algorithm is used to separate spatially distant point sets within the same cluster.
To ensure completeness, we discard boxes derived from ${\hat{n}^{3D}_{q}}$ that contain fewer than 300 and 500 points for ScanNet200 and ARKitScenes, respectively, or have volumes exceeding $8.5,\text{m}^3$, as such boxes are unlikely to represent entire objects.

\noindent\textbf{Model Training.} 
We follow the general configuration of \cite{tu2023imgeonet} to train {\ourmethod}. The 2D feature encoder of the input images $I_t$ is a ResNet-50~\cite{resnet} pretrained on ImageNet~\cite{imagenet}. In Voxel-Semantic feature alignment, we add an MLP layer atop the voxel feature to match the CLIP feature dimension. Our network is trained using AdamW~\cite{adamw} optimizer with an initial learning rate as \mdineq{1e^{-3}}. Learning rate decay is applied at the $18^{th}$ and $45^{th}$ epochs with a decay rate of 0.1, and the network undergoes 50 training epochs. 
\subsection{3D Pseudo Boxes}
\label{sec:exp-pseudo}

\begin{table}[t]
\vspace{-1em}
\small
\centering
  \caption{
  \textbf{3D Pseudo Box Evaluation on ScanNet200 and ARKitScenes.}
Our boxes, with and without Mesh Segmentation Refinement (MSR), exceed OV-3DET and SAM3D in precision at both IoU thresholds at 0.25(@25) and 0.5(@50) across both datasets. Our bounding boxes outperform OV-3DET in recall significantly and demonstrate competitive performance to SAM3D in most settings.
  (a) Please refer to the supplementary material for detailed evaluations in different subsets (head, common, tail) on ScanNet200.
  (b) 
  The $^*$ indicates that precision is expected to be low since only 17 classes are labeled in ARKitScenes. Many pseudo boxes are associated with unlabeled objects and counted as false positives.
}
\vspace{-1em}
    \resizebox{1\columnwidth}{!}{\begin{tabular}{@{}l|cccc|cccc@{}}
    \toprule
    & \multicolumn{4}{|c|}{(a) ScanNet200} & \multicolumn{4}{c}{(b) ARKitScenes} \\
    \toprule
    \multirowcell{2}[-0.6ex][l]{Method} & \multicolumn{2}{c}{Precision (\%)} & \multicolumn{2}{c|}{Recall (\%) } & \multicolumn{2}{c}{Precision$^*$ (\%)} & \multicolumn{2}{c}{Recall (\%) } \\
    \cmidrule(lr){2-3} \cmidrule(lr){4-5} \cmidrule(lr){6-7} \cmidrule(lr){8-9}
    & @25 & @50 & @25 & @50 & @25 & @50 & @25 & @50\\
    \midrule
    OV-3DET~\cite{lu2023openvocabulary} & 11.62 & 4.40 & 21.13 & 7.99 & 3.74 & 0.91 & 32.43 & 7.93 \\
    SAM3D~\cite{yang2023sam3d} & 14.48 & 9.05 & 57.70 & \textbf{36.07} & 6.01 & 1.49 & 43.78 & 10.87 \\
    \midrule
    Ours w/o MSR & 27.09 & 11.98 & 52.43 & 23.18 & \textbf{6.06} & 1.34 & 51.40 & 11.41 \\
    Ours & \textbf{32.07} & \textbf{18.14} & \textbf{58.30} & 32.99 & 5.97 & \textbf{1.58} & \textbf{51.92} & \textbf{13.74}\\
    \bottomrule
    \end{tabular}}
    \vspace{-1.5em}
  \label{table:scannet200-pseudo-box}%
\end{table}%
We evaluate our class-agnostic pseudo boxes by comparing them with ground truth boxes using various IoU thresholds. Additionally, we investigate the impact of Mesh Segmentation Refinement (MSR) on our method. The evaluation results for ScanNet200 and ARKitScenes are shown in Table~\ref{table:scannet200-pseudo-box}.
Our bounding boxes demonstrate higher quality compared to OV-3DET and SAM3D in terms of precision at IoU@0.25 and IoU@0.5. For training the detector, pseudo box precision is relatively more important than recall. 
Our bounding boxes also outperform OV-3DET in recall by a significant margin and remain comparable to SAM3D in most of the settings.
This suggests that our pseudo boxes can more effectively capture objects from various viewpoints.
OV-3DET generates pseudo boxes by back-projecting results from Detic~\cite{Detic}, an OV-2Ddet, into 3D for each 2D frame. This inherent difference between 2D and 3D domains causes OV-3DET to fall short in overall precision and recall compared to our multi-view-aware 3D pseudo box generation method.
Unlike SAM3D, which can be affected by 2D segmentation errors due to its local adjacent frame merging, {\ourmethod} utilizes graph embedding-based clustering to consider frames from all viewpoints simultaneously. This approach reduces the impact of segmentation errors from individual frames, resulting in better pseudo box quality.
In conclusion, MSR improves performance on both ScanNet200 and ARKitScenes, though the gains are linked to mesh quality, with a less pronounced effect on ARKitScenes due to its lower mesh quality. These findings indicate that {\ourmethod} consistently surpasses OV-3DET and SAM3D in pseudo box quality, irrespective of the mesh quality.
\subsection{OV 3D Object Detection Results}
\label{sec:exp-ov}

\begin{table}[t]
 \vspace{-1em}
    \begin{minipage}[t]{0.49\textwidth}
    \small
    \centering
    \caption{
      \textbf{Class-agnostic 3D Object Detection on ScanNet200.} Our proposed 3D pseudo boxes enable {\ourmethod} to achieve higher AP and AR than boxes from OV-3DET and SAM3D.
    }
    \vspace{-1em}
    \resizebox{\columnwidth}{!}{\begin{tabular}{l|c|cc}
        \toprule
        Method & Trained Box & AP@25(\%) & AR@25(\%) \\
        \midrule
        \multirow{3}{*}{\ourmethod} \
         & OV-3DET~\cite{lu2023openvocabulary} & 19.53 & 35.19\\
         & SAM3D~\cite{yang2023sam3d} & 23.77 & 47.82\\
         & Ours (w/o MSR) & 25.95 & 48.14\\
         & Ours & \textbf{26.92} & \textbf{51.19}\\
        \bottomrule
    \end{tabular}%
    }
    \label{table:agnostic-scannet200-result}
    \end{minipage}%
\hfill
  \begin{minipage}[t]{0.49\textwidth}
    \centering
    \small
    \caption{\textbf{3D Object Detection on ScanNet200.} \peng{For OpenM3D, different ``Trained Boxes'' are used in training, while for S2D, different ``Candidate Boxes'' are applied. Specifically, ``S2D+Ours'' and ``S2D+Depth Estimated" correspond to ``Baseline-S2D'' and ``Baseline-S2D using Depth Estimated 3D Boxes'', respectively.}}
    \vspace{-1em}
    \resizebox{\textwidth}{!}{
      \begin{tabular}{l|c|cc}
        \toprule
        Method & Trained Box / Candidate box & mAP@25(\%) & mAR@25(\%) \\
        \midrule
        \multirow{3}{*}{\ourmethod} 
        & OV-3DET~\cite{lu2023openvocabulary} & 3.13 & 10.83 \\
        & SAM3D~\cite{yang2023sam3d} & 3.92 & 13.33 \\
        & Ours (w/o MSR) & 4.04 & 13.77\\
        & Ours & \textbf{4.23} & \textbf{15.12}\\
        \midrule
        \multirow{2}{*}{S2D} 
        & Depth estimated & 3.80 & 8.60\\
        & Ours & 4.17 & 10.05 \\
        \bottomrule
      \end{tabular}%
    }
    \label{table:scannet200-result}
  \end{minipage}%
  \vspace{-1em}
\end{table}

\begin{table}[t]
    \begin{minipage}[t]{0.49\textwidth}
    \small
    \centering
    \caption{
      \textbf{3D Object Detection on ScanNetv2.} {\ourmethod} performs comparably to point-cloud-based methods. This table serves as a reference for comparing different input modalities in inference, including point clouds (pc) and images (im).  {\textdagger} indicates methods evaluated with OV-3DET's pseudo-boxes, while our evaluation uses ground-truth 3D boxes from ScanNetv2 in our multi-view setting.
    }
    \vspace{-1em}
    \resizebox{\columnwidth}{!}{\begin{tabular}{l|c|c|c|c}
        \toprule
        Method & Training Data & Input & Detector & AP@25(\%) \\
        \midrule
        OV-3DET{\textdagger}~\cite{lu2023openvocabulary} \
         & ScanNet & pc + im & Two-Stage & 18.02\\
        CoDA{\textdagger}~\cite{cao2023coda} \
         & ScanNet & pc & One-Stage & 19.32\\
        ImOV3D{\textdagger}~\cite{yang2024imov3d} \
         & ScanNet, LVIS & pc & One-Stage & 21.45\\
         \midrule
        Ours \
         & ScanNet & im & One-Stage & 19.76\\
        \bottomrule
    \end{tabular}%
    }
    \label{table:scannetv2-pc-im}
    \end{minipage}%
    \vspace{-1em}
\end{table}

\noindent\textbf{Class-Agnostic Scenario.}
This scenario validates the class-agnostic 3D object detector of {\ourmethod} in Sec.~\ref{sec.m-CA3D}. No class information is utilized during inference, focusing solely on accurately predicting foreground bounding boxes. We show the evaluation results of ScanNet200 in Table~\ref{table:agnostic-scannet200-result}. Compared to OV-3DET~\cite{lu2023openvocabulary}, our proposed framework can improve AP@25 by 37\% (19.53\%$\rightarrow$26.92\%), underscoring the superiority of our pseudo boxes. Given that OV-3DET generates 3D boxes solely based on a single-view RGB image and depth map, there is a risk that the resulting 3D box may deviate significantly from the actual object, thereby influencing the class-agnostic training. \ke{Additionally, we compare our framework to SAM3D, which requires multi-view frames as input. Our proposed approach consistently outperforms SAM3D by 13\% in AP@25. Furthermore, significant improvements are observed in mAR@25, with our framework surpassing SAM3D by 3.3\% and OV-3DET by 16\%.}

\noindent\textbf{Open-Vocabulary Scenario.}
This scenario validates {\ourmethod}, covering both Sec.~\ref{sec.m-CA3D} and Sec.~\ref{sec.m-align}. We evaluated OV 3D detection on ScanNet200 and report in Table~\ref{table:scannet200-result}.
Similar to the trend in Table~\ref{table:agnostic-scannet200-result}, {\ourmethod} outperforms our method trained with OV-3DET boxes on both mAP@25 and mAR@25, for 1.1\% and 4.3\%, respectively, with relative improvements exceeding 30\% on the challenging ScanNet200 dataset. \ke{Moreover, {\ourmethod} surpasses models trained with SAM3D pseudo boxes, thereby highlighting the effectiveness of our graph-embedding-based pseudo boxes. Notably, using better segmentation models, \eg, CropFormer~\cite{qilu2023high}, {\ourmethod} achieves a significant 12.5\% improvement in mAP@25, from 4.23\% to 4.76\%. For more details, please refer to the supplementary materials.} 

To highlight the contribution of our Voxel-Semantic feature alignment, we compare {\ourmethod} to \textit{S2D}, both utilizing the same class-agnostic foreground detector. As a single-stage detector, our method achieves comparable mAP@25 to \textit{S2D}. However, \textit{S2D} shows a significant drop in mAR@25, from 0.15 to 0.10, indicating that 3D voxel features recall object classes better than multi-view 2D CLIP features. Furthermore, \textit{S2D} requires the CLIP image encoder during inference, introducing high computational costs and a sixfold increase in inference time compared to our framework. On ARKitScenes, our method achieves 42.77 mAR@25, outperforming \textit{S2D} at 19.58 mAR@25, with mAP not reported for the same reason as in 3D pseudo box evaluation.

Table~\ref{table:scannetv2-pc-im} presents the results of {\ourmethod} on ScanNetv2, demonstrating performance comparable to other point-cloud-based methods. This indicates that {\ourmethod}, using only 2D images at inference, achieves results on par with methods that rely on 3D data. For additional baseline results on ScanNetv2, please refer to the supplementary material.

\begin{figure}[t]
\vspace{-1em}
\begin{center}
  \includegraphics[width=0.5\textwidth]{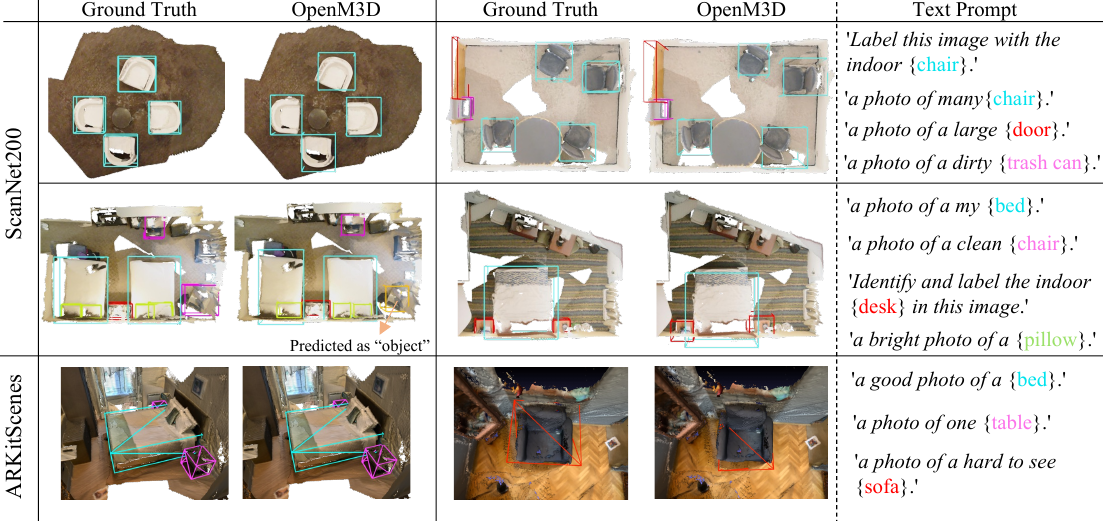}
\end{center}
\vspace{-1.5em}
\caption{
  \textbf{Qualitative Results of {\ourmethod}} on ScanNet200 and ARKitScenes. 
  Given multi-view images and corresponding camera poses, {\ourmethod} can detect objects by arbitrary text prompts towards open-vocabulary detection. The color-coded boxes correspond to different object classes. We show a subset of text prompts used in the ImageNet dataset and \textit{specific prompts}.}
\label{fig:qualitative}
\vspace{-1em}
\end{figure}

Detection results on ScanNet200 and ARKitScenes are shown in Fig.~\ref{fig:qualitative}. 
To showcase OV detection ability, we visualized detection results using a subset of text prompts by CLIP on ImageNet, and specific prompts in Fig.~\ref{fig:qualitative}.  {\ourmethod} consistently detects 3D objects across various classes using general prompts like `a photo of a large \{\}', and accurately locates specific objects, such as chairs and small desks, demonstrating its strength in OV 3D object detection.


\subsection{Ablation study}
\label{sec:exp-ablation}


\begin{figure}[t]
\centering
\includegraphics[width=1.0\columnwidth]{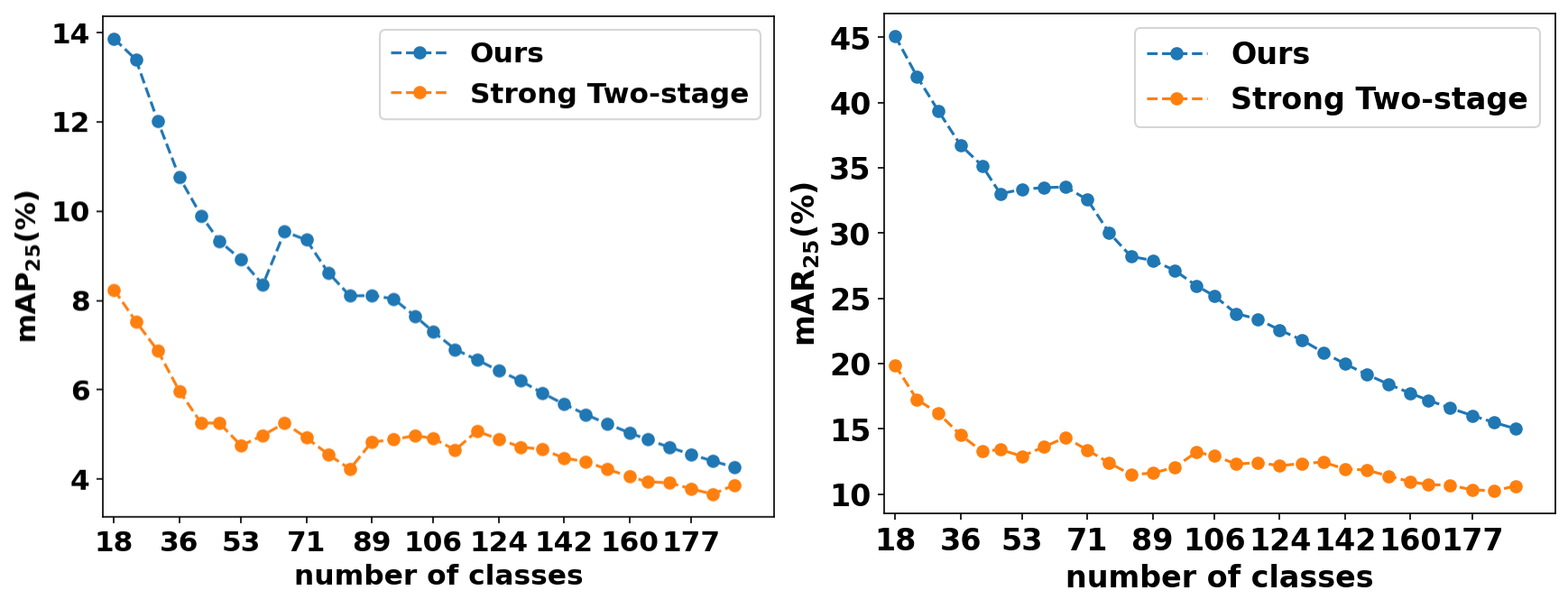}
\vspace{-2em}
\caption{\textbf{mAP{\boldmath$_{25}$} and mAR{\boldmath$_{25}$} in various class numbers from 18 to 189 classes.} As the class number gets larger, {\ourmethod} scores consistently higher mAP (\emph{Left}) and mAR (\emph{Right}) compared with the ``Strong Two-stage'' method.}
\label{fig:ablation_class_sets_ap}
\vspace{-2em}
\end{figure}


\noindent\textbf{Influence of Class Number.}
To investigate the impact of class number, we evaluate {\ourmethod} across various numbers of classes (18 to 189), following the ``head,'' ``common,'' and ``tail'' splits from ScanNet200. As shown in Fig.~\ref{fig:ablation_class_sets_ap}, our method consistently achieves higher mAP and mAR than the ``Strong Two-stage'' method as the class number increases.

\noindent\textbf{Different Prompts.} 
CLIP~\cite{CLIP} indicated that multiple prompts benefit more comprehensive understanding of the desired context, and provided a list of prompts for various datasets depending on the dataset domains. {\ourmethod} predicts classes by matching the 3D voxel feature with the text embeddings of class names wrapped in multiple prompts, such as ``\texttt{A photo of a \{\}}'', and selects the category with the highest cosine similarity. We apply the same here and mostly used the prompts of ImageNet~\cite{imagenet} dataset. We primarily used ImageNet~\cite{imagenet} prompts but also evaluated our model with Cifar100~\cite{krizhevsky2009learning} prompts. The Cifar100 prompts resulted in mAP@25 and mAR@25 values of 4.14 and 14.79, respectively, showing minimal difference in performance . 
This implies that adjusting the prompts might not significantly improve performance.
\section{Conclusion}
\label{sec:conclusion}
We introduced {\ourmethod}, a novel single-stage open-vocabulary multi-view 3D object detector trained without human annotations. It leverages class-agnostic 3D localization and voxel-semantic alignment losses, guided by high-quality 3D pseudo-boxes and diverse CLIP features. We introduce a graph-based 3D pseudo-box generation method achieving superior precision and recall in pseudo-box quality than OV-3DET and SAM3D. At inference, requiring only multi-view images, {\ourmethod} outperforms a strong two-stage approach and models trained with OV-3DET and SAM3D boxes on ScanNet200 and ARKitScenes, while excelling over an estimated multi-view depth baseline in accuracy and speed.




\begin{thebibliography}{66}
\providecommand{\natexlab}[1]{#1}
\providecommand{\url}[1]{\texttt{#1}}
\expandafter\ifx\csname urlstyle\endcsname\relax
  \providecommand{\doi}[1]{doi: #1}\else
  \providecommand{\doi}{doi: \begingroup \urlstyle{rm}\Url}\fi

\bibitem[Bae et~al.(2022)Bae, Budvytis, and Cipolla]{Bae2022}
Gwangbin Bae, Ignas Budvytis, and Roberto Cipolla.
\newblock Multi-view depth estimation by fusing single-view depth probability with multi-view geometry.
\newblock In \emph{Proc. IEEE Conference on Computer Vision and Pattern Recognition (CVPR)}, 2022.

\bibitem[Baruch et~al.(2021)Baruch, Chen, Dehghan, Dimry, Feigin, Fu, Gebauer, Joffe, Kurz, Schwartz, and Shulman]{arkitscenes}
Gilad Baruch, Zhuoyuan Chen, Afshin Dehghan, Tal Dimry, Yuri Feigin, Peter Fu, Thomas Gebauer, Brandon Joffe, Daniel Kurz, Arik Schwartz, and Elad Shulman.
\newblock Arkitscenes - a diverse real-world dataset for 3d indoor scene understanding using mobile rgb-d data.
\newblock In \emph{Advances in Neural Information Processing Systems (NeurIPS)}, 2021.

\bibitem[Cao et~al.(2023)Cao, Zeng, Xu, and Xu]{cao2023coda}
Yang Cao, Yihan Zeng, Hang Xu, and Dan Xu.
\newblock Coda: Collaborative novel box discovery and cross-modal alignment for open-vocabulary 3d object detection.
\newblock In \emph{Advances in Neural Information Processing Systems (NeurIPS)}, 2023.

\bibitem[Carion et~al.(2020)Carion, Massa, Synnaeve, Usunier, Kirillov, and Zagoruyko]{carion2020detr}
Nicolas Carion, Francisco Massa, Gabriel Synnaeve, Nicolas Usunier, Alexander Kirillov, and Sergey Zagoruyko.
\newblock End-to-end object detection with transformers.
\newblock In \emph{European Conference on Computer Vision (ECCV)}, 2020.

\bibitem[Dai et~al.(2017)Dai, Chang, Savva, Halber, Funkhouser, and Nie{\ss}ner]{dai2017scannet}
Angela Dai, Angel~X. Chang, Manolis Savva, Maciej Halber, Thomas Funkhouser, and Matthias Nie{\ss}ner.
\newblock Scannet: Richly-annotated 3d reconstructions of indoor scenes.
\newblock In \emph{IEEE Conference on Computer Vision and Pattern Recognition (CVPR)}, 2017.

\bibitem[Deng et~al.(2009)Deng, Dong, Socher, Li, Li, and Fei-Fei]{imagenet}
Jia Deng, Wei Dong, Richard Socher, Li-Jia Li, Kai Li, and Li Fei-Fei.
\newblock Imagenet: A large-scale hierarchical image database.
\newblock In \emph{IEEE Conference on Computer Vision and Pattern Recognition (CVPR)}, 2009.

\bibitem[Deng et~al.(2021)Deng, Shi, Li, Zhou, Zhang, and Li]{deng2021voxelrcnn}
Jiajun Deng, Shaoshuai Shi, Peiwei Li, Wengang Zhou, Yanyong Zhang, and Houqiang Li.
\newblock Voxel r-cnn: Towards high performance voxel-based 3d object detection.
\newblock In \emph{AAAI Conference on Artificial Intelligence (AAAI)}, 2021.

\bibitem[Felzenszwalb and Huttenlocher(2004)]{felzenszwalb2004efficient}
Pedro~F Felzenszwalb and Daniel~P Huttenlocher.
\newblock Efficient graph-based image segmentation.
\newblock \emph{International Journal of Computer Vision (IJCV)}, 59, 2004.

\bibitem[Ghiasi et~al.(2022{\natexlab{a}})Ghiasi, Gu, Cui, and Lin]{OpenSeg}
Golnaz Ghiasi, Xiuye Gu, Yin Cui, and Tsung{-}Yi Lin.
\newblock Scaling open-vocabulary image segmentation with image-level labels.
\newblock In \emph{European Conference on Computer Vision (ECCV)}, 2022{\natexlab{a}}.

\bibitem[Ghiasi et~al.(2022{\natexlab{b}})Ghiasi, Gu, Cui, and Lin]{ghiasi2022scaling}
Golnaz Ghiasi, Xiuye Gu, Yin Cui, and Tsung-Yi Lin.
\newblock Scaling open-vocabulary image segmentation with image-level labels.
\newblock In \emph{European Conference on Computer Vision (ECCV)}, 2022{\natexlab{b}}.

\bibitem[Gu et~al.(2022)Gu, Lin, Kuo, and Cui]{gu2022openvocabulary}
Xiuye Gu, Tsung-Yi Lin, Weicheng Kuo, and Yin Cui.
\newblock Open-vocabulary object detection via vision and language knowledge distillation.
\newblock In \emph{International Conference on Learning Representations (ICLR)}, 2022.

\bibitem[Gwak et~al.(2020)Gwak, Choy, and Savarese]{gwak2020gsdn}
JunYoung Gwak, Christopher Choy, and Silvio Savarese.
\newblock Generative sparse detection networks for 3d single-shot object detection.
\newblock In \emph{European Conference on Computer Vision (ECCV)}, 2020.

\bibitem[Ha and Song(2022)]{ha2022semantic}
Huy Ha and Shuran Song.
\newblock Semantic abstraction: Open-world 3d scene understanding from 2d vision-language models.
\newblock In \emph{Conference on Robot Learning (CoRL)}, 2022.

\bibitem[He et~al.(2016)He, Zhang, Ren, and Sun]{resnet}
Kaiming He, Xiangyu Zhang, Shaoqing Ren, and Jian Sun.
\newblock Deep residual learning for image recognition.
\newblock In \emph{IEEE Conference on Computer Vision and Pattern Recognition (CVPR)}, 2016.

\bibitem[Huang et~al.(2021)Huang, Huang, Zhu, and Du]{huang2021bevdet}
Junjie Huang, Guan Huang, Zheng Zhu, and Dalong Du.
\newblock Bevdet: High-performance multi-camera 3d object detection in bird-eye-view.
\newblock \emph{arXiv preprint arXiv:2112.11790}, 2021.

\bibitem[Jatavallabhula et~al.(2023)Jatavallabhula, Kuwajerwala, Gu, Omama, Chen, Li, Iyer, Saryazdi, Keetha, Tewari, Tenenbaum, {de Melo}, Krishna, Paull, Shkurti, and Torralba]{conceptfusion}
{Krishna Murthy} Jatavallabhula, Alihusein Kuwajerwala, Qiao Gu, Mohd Omama, Tao Chen, Shuang Li, Ganesh Iyer, Soroush Saryazdi, Nikhil Keetha, Ayush Tewari, {Joshua B.} Tenenbaum, {Celso Miguel} {de Melo}, Madhava Krishna, Liam Paull, Florian Shkurti, and Antonio Torralba.
\newblock Conceptfusion: Open-set multimodal 3d mapping.
\newblock In \emph{Robotics: Science and Systems (RSS)}, 2023.

\bibitem[Jia et~al.(2021)Jia, Yang, Xia, Chen, Parekh, Pham, Le, Sung, Li, and Duerig]{ALIGN}
Chao Jia, Yinfei Yang, Ye Xia, Yi-Ting Chen, Zarana Parekh, Hieu Pham, Quoc Le, Yun-Hsuan Sung, Zhen Li, and Tom Duerig.
\newblock Scaling up visual and vision-language representation learning with noisy text supervision.
\newblock In \emph{International Conference on Machine Learning (ICML)}, 2021.

\bibitem[Kerr et~al.(2023)Kerr, Kim, Goldberg, Kanazawa, and Tancik]{kerr2023lerf}
Justin Kerr, Chung~Min Kim, Ken Goldberg, Angjoo Kanazawa, and Matthew Tancik.
\newblock Lerf: Language embedded radiance fields.
\newblock In \emph{Proceedings of the IEEE/CVF International Conference on Computer Vision}, pages 19729--19739, 2023.

\bibitem[Kirillov et~al.(2023)Kirillov, Mintun, Ravi, Mao, Rolland, Gustafson, Xiao, Whitehead, Berg, Lo, Dollar, and Girshick]{SAM}
Alexander Kirillov, Eric Mintun, Nikhila Ravi, Hanzi Mao, Chloe Rolland, Laura Gustafson, Tete Xiao, Spencer Whitehead, Alexander~C. Berg, Wan-Yen Lo, Piotr Dollar, and Ross Girshick.
\newblock Segment anything.
\newblock In \emph{IEEE International Conference on Computer Vision (ICCV)}, 2023.

\bibitem[Krizhevsky et~al.(2009)Krizhevsky, Hinton, et~al.]{krizhevsky2009learning}
Alex Krizhevsky, Geoffrey Hinton, et~al.
\newblock Learning multiple layers of features from tiny images.
\newblock 2009.

\bibitem[Lang et~al.(2019)Lang, Vora, Caesar, Zhou, Yang, and Beijbom]{lang2019pointpillars}
Alex~H Lang, Sourabh Vora, Holger Caesar, Lubing Zhou, Jiong Yang, and Oscar Beijbom.
\newblock Pointpillars: Fast encoders for object detection from point clouds.
\newblock In \emph{IEEE Conference on Computer Vision and Pattern Recognition (CVPR)}, 2019.

\bibitem[Li et~al.(2023)Li, Zhang, Xu, Liu, Zhang, Ni, and Shum]{li2022mask}
Feng Li, Hao Zhang, Huaizhe Xu, Shilong Liu, Lei Zhang, Lionel~M Ni, and Heung-Yeung Shum.
\newblock Mask dino: Towards a unified transformer-based framework for object detection and segmentation.
\newblock In \emph{IEEE Conference on Computer Vision and Pattern Recognition (CVPR)}, 2023.

\bibitem[Li et~al.(2022)Li, Wang, Li, Xie, Sima, Lu, Qiao, and Dai]{li2022bevformer}
Zhiqi Li, Wenhai Wang, Hongyang Li, Enze Xie, Chonghao Sima, Tong Lu, Yu Qiao, and Jifeng Dai.
\newblock Bevformer: Learning bird’s-eye-view representation from multi-camera images via spatiotemporal transformers.
\newblock In \emph{European Conference on Computer Vision (ECCV)}, 2022.

\bibitem[Lin et~al.(2017)Lin, Goyal, Girshick, He, and Doll{\'a}r]{lin2017focal}
Tsung-Yi Lin, Priya Goyal, Ross Girshick, Kaiming He, and Piotr Doll{\'a}r.
\newblock Focal loss for dense object detection.
\newblock In \emph{IEEE International Conference on Computer Vision (ICCV)}, 2017.

\bibitem[Liu et~al.(2022)Liu, Wang, Zhang, and Sun]{liu2022petr}
Yingfei Liu, Tiancai Wang, Xiangyu Zhang, and Jian Sun.
\newblock Petr: Position embedding transformation for multi-view 3d object detection.
\newblock In \emph{European Conference on Computer Vision (ECCV)}, 2022.

\bibitem[Loshchilov and Hutter(2019)]{adamw}
Ilya Loshchilov and Frank Hutter.
\newblock Decoupled weight decay regularization.
\newblock In \emph{International Conference on Learning Representations (ICLR)}, 2019.

\bibitem[Lu et~al.(2023{\natexlab{a}})Lu, Kuen, Tiancheng, Jiuxiang, Weidong, Jiaya, Zhe, and Ming-Hsuan]{qilu2023high}
Qi Lu, Jason Kuen, Shen Tiancheng, Gu Jiuxiang, Guo Weidong, Jia Jiaya, Lin Zhe, and Yang Ming-Hsuan.
\newblock High-quality entity segmentation.
\newblock In \emph{IEEE International Conference on Computer Vision (ICCV)}, 2023{\natexlab{a}}.

\bibitem[Lu et~al.(2023{\natexlab{b}})Lu, Chang, Jing, Boularias, and Bekris]{lu2023ovir}
Shiyang Lu, Haonan Chang, Eric~Pu Jing, Abdeslam Boularias, and Kostas Bekris.
\newblock Ovir-3d: Open-vocabulary 3d instance retrieval without training on 3d data.
\newblock In \emph{7th Annual Conference on Robot Learning}, 2023{\natexlab{b}}.

\bibitem[Lu et~al.(2023{\natexlab{c}})Lu, Xu, Wei, Xie, Tomizuka, Keutzer, and Zhang]{lu2023openvocabulary}
Yuheng Lu, Chenfeng Xu, Xiaobao Wei, Xiaodong Xie, Masayoshi Tomizuka, Kurt Keutzer, and Shanghang Zhang.
\newblock Open-vocabulary point-cloud object detection without 3d annotation.
\newblock In \emph{IEEE Conference on Computer Vision and Pattern Recognition (CVPR)}, 2023{\natexlab{c}}.

\bibitem[Mao et~al.(2021)Mao, Xue, Niu, Bai, Feng, Liang, Xu, and Xu]{mao2021voxeltransformer}
Jiageng Mao, Yujing Xue, Minzhe Niu, Haoyue Bai, Jiashi Feng, Xiaodan Liang, Hang Xu, and Chunjing Xu.
\newblock Voxel transformer for 3d object detection.
\newblock In \emph{IEEE International Conference on Computer Vision (ICCV)}, 2021.

\bibitem[Murez et~al.(2020)Murez, Van~As, Bartolozzi, Sinha, Badrinarayanan, and Rabinovich]{murez2020atlas}
Zak Murez, Tarrence Van~As, James Bartolozzi, Ayan Sinha, Vijay Badrinarayanan, and Andrew Rabinovich.
\newblock Atlas: End-to-end 3d scene reconstruction from posed images.
\newblock In \emph{European Conference on Computer Vision (ECCV)}, 2020.

\bibitem[Pan et~al.(2021)Pan, Xia, Song, Li, and Huang]{pan2021pointformer}
Xuran Pan, Zhuofan Xia, Shiji Song, Li~Erran Li, and Gao Huang.
\newblock 3d object detection with pointformer.
\newblock In \emph{IEEE Conference on Computer Vision and Pattern Recognition (CVPR)}, 2021.

\bibitem[Peng et~al.(2023)Peng, Genova, Jiang, Tagliasacchi, Pollefeys, and Funkhouser]{Peng2023OpenScene}
Songyou Peng, Kyle Genova, Chiyu~"Max" Jiang, Andrea Tagliasacchi, Marc Pollefeys, and Thomas Funkhouser.
\newblock Openscene: 3d scene understanding with open vocabularies.
\newblock In \emph{IEEE Conference on Computer Vision and Pattern Recognition (CVPR)}, 2023.

\bibitem[Perozzi et~al.(2014)Perozzi, Al-Rfou, and Skiena]{perozzi2014deepwalk}
Bryan Perozzi, Rami Al-Rfou, and Steven Skiena.
\newblock Deepwalk: Online learning of social representations.
\newblock In \emph{Proceedings of the 20th ACM SIGKDD international conference on Knowledge discovery and data mining}, pages 701--710, 2014.

\bibitem[Qi et~al.(2017)Qi, Yi, Su, and Guibas]{qi2017pointnet++}
Charles~Ruizhongtai Qi, Li Yi, Hao Su, and Leonidas~J Guibas.
\newblock Pointnet++: Deep hierarchical feature learning on point sets in a metric space.
\newblock In \emph{Advances in Neural Information Processing Systems (NeurIPS)}, 2017.

\bibitem[Qi et~al.(2018)Qi, Liu, Wu, Su, and Guibas]{qi2018fpointnet}
Charles~R Qi, Wei Liu, Chenxia Wu, Hao Su, and Leonidas~J Guibas.
\newblock Frustum pointnets for 3d object detection from rgb-d data.
\newblock In \emph{IEEE Conference on Computer Vision and Pattern Recognition (CVPR)}, 2018.

\bibitem[Qi et~al.(2019)Qi, Litany, He, and Guibas]{qi2019votenet}
Charles~R Qi, Or Litany, Kaiming He, and Leonidas~J Guibas.
\newblock Deep hough voting for 3d object detection in point clouds.
\newblock In \emph{IEEE International Conference on Computer Vision (ICCV)}, 2019.

\bibitem[Radford et~al.(2021)Radford, Kim, Hallacy, Ramesh, Goh, Agarwal, Sastry, Askell, Mishkin, Clark, Krueger, and Sutskever]{CLIP}
Alec Radford, Jong~Wook Kim, Chris Hallacy, Aditya Ramesh, Gabriel Goh, Sandhini Agarwal, Girish Sastry, Amanda Askell, Pamela Mishkin, Jack Clark, Gretchen Krueger, and Ilya Sutskever.
\newblock Learning transferable visual models from natural language supervision.
\newblock In \emph{International Conference on Machine Learning (ICML)}, 2021.

\bibitem[Rahman et~al.(2020)Rahman, Khan, and Barnes]{RahmanAAAI20}
Shafin Rahman, Salman Khan, and Nick Barnes.
\newblock Improved visual-semantic alignment for zero-shot object detection.
\newblock In \emph{AAAI Conference on Artificial Intelligence (AAAI)}, 2020.

\bibitem[Rozenberszki et~al.(2022)Rozenberszki, Litany, and Dai]{rozenberszki2022language}
David Rozenberszki, Or Litany, and Angela Dai.
\newblock Language-grounded indoor 3d semantic segmentation in the wild.
\newblock In \emph{European Conference on Computer Vision (ECCV)}, 2022.

\bibitem[Rukhovich et~al.(2022{\natexlab{a}})Rukhovich, Vorontsova, and Konushin]{rukhovich2022fcaf3d}
Danila Rukhovich, Anna Vorontsova, and Anton Konushin.
\newblock Fcaf3d: fully convolutional anchor-free 3d object detection.
\newblock In \emph{European Conference on Computer Vision (ECCV)}, 2022{\natexlab{a}}.

\bibitem[Rukhovich et~al.(2022{\natexlab{b}})Rukhovich, Vorontsova, and Konushin]{rukhovich2022imvoxelnet}
Danila Rukhovich, Anna Vorontsova, and Anton Konushin.
\newblock Imvoxelnet: Image to voxels projection for monocular and multi-view general-purpose 3d object detection.
\newblock In \emph{Winter Conference on Applications of Computer Vision (WACV)}, 2022{\natexlab{b}}.

\bibitem[Shi et~al.(2019)Shi, Wang, and Li]{shi2019pointrcnn}
Shaoshuai Shi, Xiaogang Wang, and Hongsheng Li.
\newblock Pointrcnn: 3d object proposal generation and detection from point cloud.
\newblock In \emph{IEEE Conference on Computer Vision and Pattern Recognition (CVPR)}, 2019.

\bibitem[Shi et~al.(2020)Shi, Wang, Shi, Wang, and Li]{shi2020pillarod}
Shaoshuai Shi, Zhe Wang, Jianping Shi, Xiaogang Wang, and Hongsheng Li.
\newblock From points to parts: 3d object detection from point cloud with part-aware and part-aggregation network.
\newblock \emph{IEEE Transactions on Pattern Analysis and Machine Intelligence (TPAMI)}, 2020.

\bibitem[Shi and Rajkumar(2020)]{shi2020pointgnn}
Weijing Shi and Raj Rajkumar.
\newblock Point-gnn: Graph neural network for 3d object detection in a point cloud.
\newblock In \emph{IEEE Conference on Computer Vision and Pattern Recognition (CVPR)}, 2020.

\bibitem[Takmaz et~al.(2023)Takmaz, Fedele, Sumner, Pollefeys, Tombari, and Engelmann]{takmaz2023openmask3d}
Ayça Takmaz, Elisabetta Fedele, Robert~W. Sumner, Marc Pollefeys, Federico Tombari, and Francis Engelmann.
\newblock Openmask3d: Open-vocabulary 3d instance segmentation.
\newblock In \emph{Advances in Neural Information Processing Systems (NeurIPS)}, 2023.

\bibitem[Tang et~al.(2024)Tang, Fan, Wang, Xu, Ranjan, Schwing, and Yan]{tang2024mv}
Zhenggang Tang, Yuchen Fan, Dilin Wang, Hongyu Xu, Rakesh Ranjan, Alexander Schwing, and Zhicheng Yan.
\newblock Mv-dust3r+: Single-stage scene reconstruction from sparse views in 2 seconds.
\newblock \emph{arXiv preprint arXiv:2412.06974}, 2024.

\bibitem[Tian et~al.(2020)Tian, Shen, Chen, and He]{tian2020fcos}
Zhi Tian, Chunhua Shen, Hao Chen, and Tong He.
\newblock Fcos: A simple and strong anchor-free object detector.
\newblock \emph{IEEE Transactions on Pattern Analysis and Machine Intelligence (TPAMI)}, 2020.

\bibitem[Tseng et~al.(2022)Tseng, Chen, Lee, Wu, Chen, and Hsu]{tseng2022crossdtr}
Ching-Yu Tseng, Yi-Rong Chen, Hsin-Ying Lee, Tsung-Han Wu, Wen-Chin Chen, and Winston Hsu.
\newblock Crossdtr: Cross-view and depth-guided transformers for 3d object detection.
\newblock \emph{arXiv preprint arXiv:2209.13507}, 2022.

\bibitem[Tu et~al.(2023)Tu, Chuang, Liu, Sun, Zhang, Roy, Kuo, and Sun]{tu2023imgeonet}
Tao Tu, Shun-Po Chuang, Yu-Lun Liu, Cheng Sun, Ke Zhang, Donna Roy, Cheng-Hao Kuo, and Min Sun.
\newblock Imgeonet: Image-induced geometry-aware voxel representation for multi-view 3d object detection.
\newblock In \emph{IEEE International Conference on Computer Vision (ICCV)}, 2023.

\bibitem[Wang et~al.(2025)Wang, Chen, Karaev, Vedaldi, Rupprecht, and Novotny]{wang2025vggt}
Jianyuan Wang, Minghao Chen, Nikita Karaev, Andrea Vedaldi, Christian Rupprecht, and David Novotny.
\newblock Vggt: Visual geometry grounded transformer.
\newblock In \emph{Proceedings of the IEEE/CVF Conference on Computer Vision and Pattern Recognition}, 2025.

\bibitem[Wang et~al.(2022)Wang, Guizilini, Zhang, Wang, Zhao, and Solomon]{wang2022detr3d}
Yue Wang, Vitor~Campagnolo Guizilini, Tianyuan Zhang, Yilun Wang, Hang Zhao, and Justin Solomon.
\newblock Detr3d: 3d object detection from multi-view images via 3d-to-2d queries.
\newblock In \emph{Conference on Robot Learning (CoRL)}, 2022.

\bibitem[Xu et~al.(2023)Xu, Wu, Hou, Tsai, Li, Wang, Zhan, He, Vajda, Keutzer, and Tomizuka]{xu2023nerfdet}
Chenfeng Xu, Bichen Wu, Ji Hou, Sam Tsai, Ruilong Li, Jialiang Wang, Wei Zhan, Zijian He, Peter Vajda, Kurt Keutzer, and Masayoshi Tomizuka.
\newblock Nerf-det: Learning geometry-aware volumetric representation for multi-view 3d object detection.
\newblock In \emph{IEEE International Conference on Computer Vision (ICCV)}, 2023.

\bibitem[Xu et~al.(2022)Xu, Zhang, Wei, Lin, Cao, Hu, and Bai]{xu2022simple}
Mengde Xu, Zheng Zhang, Fangyun Wei, Yutong Lin, Yue Cao, Han Hu, and Xiang Bai.
\newblock A simple baseline for open-vocabulary semantic segmentation with pre-trained vision-language model.
\newblock In \emph{European Conference on Computer Vision (ECCV)}, 2022.

\bibitem[Yan et~al.(2018)Yan, Mao, and Li]{yan2018second}
Yan Yan, Yuxing Mao, and Bo Li.
\newblock Second: Sparsely embedded convolutional detection.
\newblock \emph{Sensors}, 2018.

\bibitem[Yang et~al.(2018)Yang, Luo, and Urtasun]{yang2018pixor}
Bin Yang, Wenjie Luo, and Raquel Urtasun.
\newblock Pixor: Real-time 3d object detection from point clouds.
\newblock In \emph{IEEE Conference on Computer Vision and Pattern Recognition (CVPR)}, 2018.

\bibitem[Yang et~al.(2024)Yang, Ju, and Yi]{yang2024imov3d}
Timing Yang, Yuanliang Ju, and Li Yi.
\newblock Imov3d: Learning open-vocabulary point clouds 3d object detection from only 2d images.
\newblock \emph{NeurIPS 2024}, 2024.

\bibitem[Yang et~al.(2023)Yang, Wu, He, Zhao, and Liu]{yang2023sam3d}
Yunhan Yang, Xiaoyang Wu, Tong He, Hengshuang Zhao, and Xihui Liu.
\newblock Sam3d: Segment anything in 3d scenes.
\newblock \emph{arXiv preprint arXiv:2306.03908}, 2023.

\bibitem[Yang et~al.(2019)Yang, Sun, Liu, Shen, and Jia]{yang2019std}
Zetong Yang, Yanan Sun, Shu Liu, Xiaoyong Shen, and Jiaya Jia.
\newblock Std: Sparse-to-dense 3d object detector for point cloud.
\newblock In \emph{IEEE International Conference on Computer Vision (ICCV)}, 2019.

\bibitem[Yang et~al.(2020)Yang, Sun, Liu, and Jia]{yang20203dssd}
Zetong Yang, Yanan Sun, Shu Liu, and Jiaya Jia.
\newblock 3dssd: Point-based 3d single stage object detector.
\newblock In \emph{IEEE Conference on Computer Vision and Pattern Recognition (CVPR)}, 2020.

\bibitem[Zareian et~al.(2021)Zareian, Rosa, Hu, and Chang]{ZareianCVPR21}
Alireza Zareian, Kevin~Dela Rosa, Derek~Hao Hu, and Shih-Fu Chang.
\newblock Open-vocabulary object detection using captions.
\newblock In \emph{IEEE Conference on Computer Vision and Pattern Recognition (CVPR)}, 2021.

\bibitem[Zhang et~al.(2021)Zhang, Guo, Zhang, Li, Miao, Cui, Qiao, Gao, and Li]{zhang2021pointclip}
Renrui Zhang, Ziyu Guo, Wei Zhang, Kunchang Li, Xupeng Miao, Bin Cui, Yu Qiao, Peng Gao, and Hongsheng Li.
\newblock Pointclip: Point cloud understanding by clip.
\newblock \emph{arXiv preprint arXiv:2112.02413}, 2021.

\bibitem[Zhang et~al.(2020)Zhang, Sun, Yang, and Huang]{zhang2020h3dnet}
Zaiwei Zhang, Bo Sun, Haitao Yang, and Qixing Huang.
\newblock H3dnet: 3d object detection using hybrid geometric primitives.
\newblock In \emph{European Conference on Computer Vision (ECCV)}, 2020.

\bibitem[Zhou et~al.(2019)Zhou, Fang, Song, Guan, Yin, Dai, and Yang]{zhou2019iou}
Dingfu Zhou, Jin Fang, Xibin Song, Chenye Guan, Junbo Yin, Yuchao Dai, and Ruigang Yang.
\newblock Iou loss for 2d/3d object detection.
\newblock In \emph{International Conference on 3D Vision (3DV)}, 2019.

\bibitem[Zhou et~al.(2022)Zhou, Girdhar, Joulin, Kr{\"a}henb{\"u}hl, and Misra]{Detic}
Xingyi Zhou, Rohit Girdhar, Armand Joulin, Philipp Kr{\"a}henb{\"u}hl, and Ishan Misra.
\newblock Detecting twenty-thousand classes using image-level supervision.
\newblock In \emph{European Conference on Computer Vision (ECCV)}, 2022.

\bibitem[Zhou and Tuzel(2018)]{zhou2018voxelnet}
Yin Zhou and Oncel Tuzel.
\newblock Voxelnet: End-to-end learning for point cloud based 3d object detection.
\newblock In \emph{IEEE Conference on Computer Vision and Pattern Recognition (CVPR)}, 2018.

\end{thebibliography}

\appendix
\section{Visualization}
This section showcases visualizations of 3D pseudo boxes generated by our method, along with additional qualitative results from \ourmethod.

\begin{figure*}[ht!]
\begin{center}
  \includegraphics[width=0.85\textwidth]{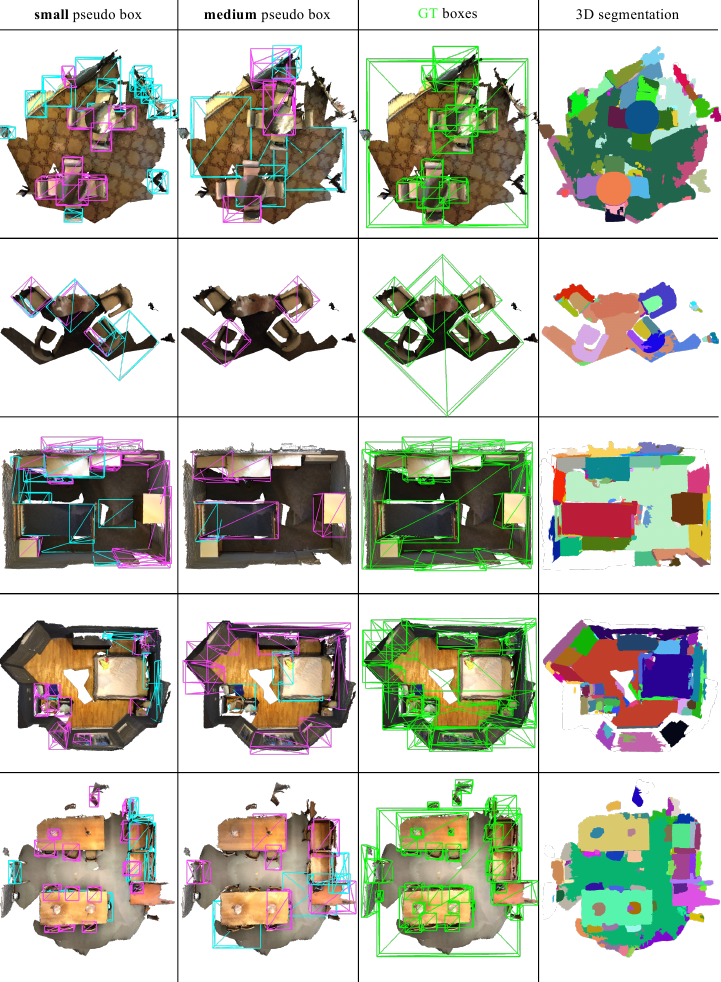}
\end{center}
\caption{
  \textbf{Visualize Pseudo Boxes of {\ourmethod}} on ScanNet200.
  We visualize our 3D pseudo boxes using two different volume sizes (small and medium). In this visualization, \textcolor{cyan}{cyan} represents false positives, while \textcolor{magenta}{magenta} represents true positives matching the \textcolor{green}{GT boxes} at IoU@0.25.}
\label{fig:pseudo_box_qualitative}
\end{figure*}

\begin{figure*}[ht!]
\begin{center}
  \includegraphics[width=\textwidth]{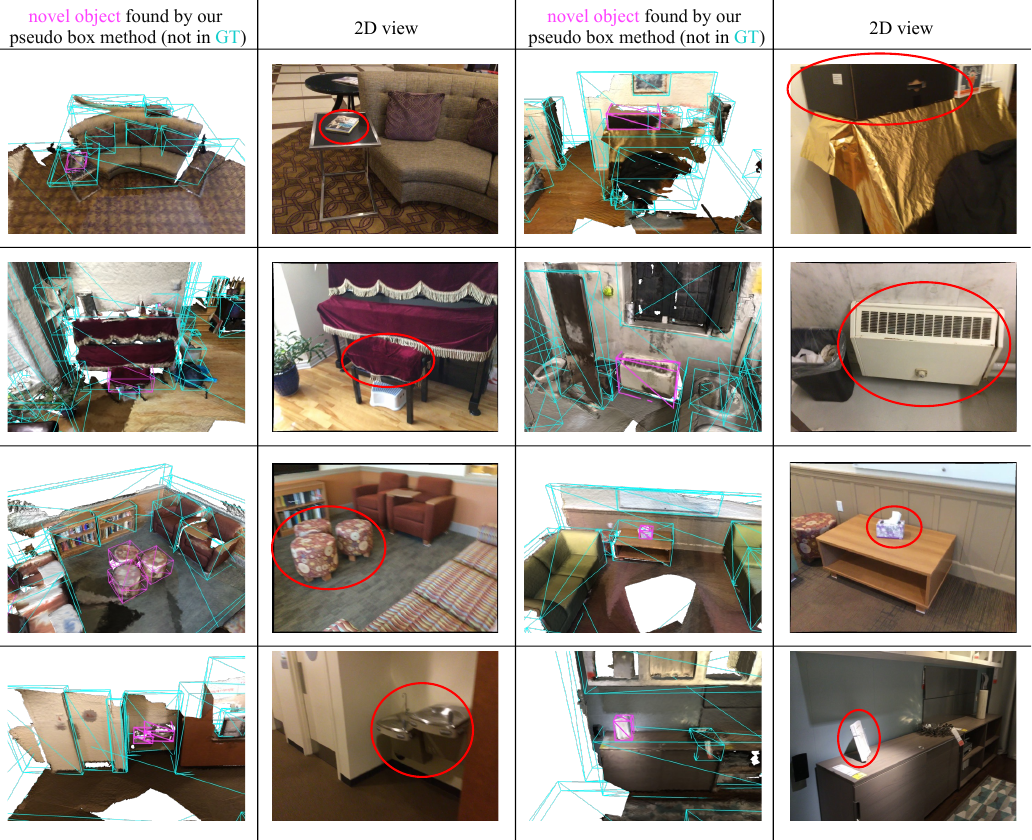}
\end{center}
\vspace{-1.5em}
\caption{
  \textbf{Localizing Novel Object with Pseudo Box} on ScanNet200.}
\label{fig:novel_pseudo}
\end{figure*}

\noindent\textbf{Visualize 3D Pseudo Boxes.}  
The localization capability of our pseudo boxes has been validated in Table \textcolor{blue}{1}, \textcolor{blue}{2} of the main paper. In Fig.~\ref{fig:pseudo_box_qualitative}, we show some examples of our 3D pseudo boxes and their corresponding 3D segmentations. To clearly present our pseudo boxes, we organize them based on two distinct ranges—small and medium—using the volumes of the boxes. 
Moreover, our 3D pseudo boxes can accurately locate novel objects, as illustrated in Fig.~\ref{fig:novel_pseudo}, in addition to those annotated in the ground truth. 
These results validate the localization capability of our generated class-agnostic pseudo boxes for various potential objects in the scene, paving the way for open-vocabulary 3D object detection.

\begin{figure*}[ht!]
\begin{center}
  \includegraphics[width=\textwidth]{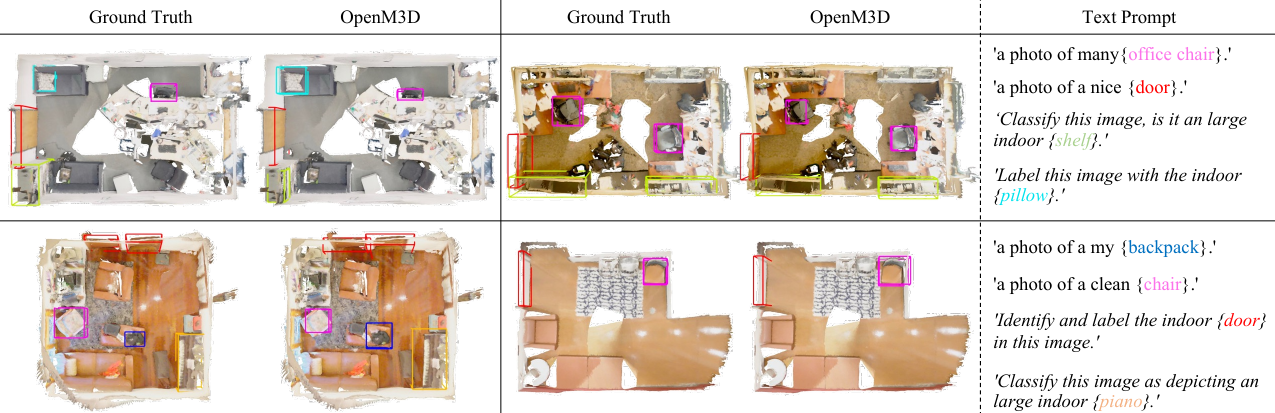}
\end{center}
\vspace{-1.2em}
\caption{
  \textbf{More Qualitative Results of {\ourmethod}} on ScanNet200.  We show general text prompts used in the ImageNet dataset, as well as prompts from \textit{specific text}. 
  }
\label{fig:more_qualitative}
\end{figure*}

\begin{figure}[ht!]
\begin{center}
  \includegraphics{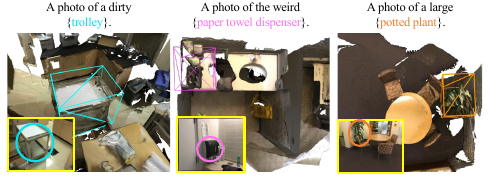}
\end{center}
\vspace{-1.2em}
\caption{
  \textbf{Novel and tail predictions in {\ourmethod}}.}
\label{fig:novel_tail}
\end{figure}

\noindent\textbf{More Qualitative Results.}
We present more qualitative results of open-vocabulary 3D object detection obtained by {\ourmethod} in Fig.~\ref{fig:more_qualitative}. Some detection results for tail and novel objects are also shown in Fig.~\ref{fig:novel_tail}.
With general prompts used in CLIP, {\ourmethod} demonstrates consistent 3D detections across multiple classes. 
This strongly showcases {\ourmethod}'s capability in open-vocabulary 3D object detection.
\section{Experiment Details}

\subsection{Implementation Details}
\noindent\textbf{Frame Selection for Generating 2D Segments. } 
To achieve fine-grained SAM~\cite{SAM} results for each frame, we aim to automatically choose frames with distinct outlines as SAM inputs for improving segmentation quality. As a result, we utilize Laplacian calculations to determine sharpness as the basis for selecting frames. For every scene in ScanNet200 and ARKitScenes, we divide all frames into intervals based on chronological order. Within each interval, we select the frame with the highest sharpness value. This process was repeated until 300 frames were chosen, iterating through the remaining frames in each round.

\noindent\textbf{2D Segments Filtering and Refinement.} 
When generating 2D segments from an image, we noticed that SAM may generate excessively small segments. Such problematic segments confuse the CLIP image encoder, resulting in poor embedding quality. Multiple small segments may map to the same voxel and further worsen the open-vocabulary classification of {\ourmethod}. We thus add preprocessing steps to exclude such patches: We set a minimum bounding-box size of 30 pixels, and a 0.02 ratio threshold of observed 3D points within each segment’s bounding box. 

Given that CLIP is trained using real-world images, our approach involves incorporating the surrounding regions of the bounding box when calculating the CLIP embedding for each 2D segment, to provide a scenario similar to real-world images. In addition to the image patch tightly cropped by the bounding box around each segment, we include patches from areas surrounding the segment with dimensions of $110\%$ and $120\%$ relative to the size of the bounding box. In OV-3DET, Lu \etal use a predefined vocabulary with 364 categories for pseudo box generation, we follow the same and use the vocabulary to improve on CLIP segment embeddings. Specifically, for each segment, we compare the embedding of the segment to the text embedding of all categories, and use the embedding of the closest category. 
Please refer to Alg.~\ref{alg:pseudo_box_gen} for the pseudo code of 3D pseudo box generation.

\begin{algorithm}
    \SetKwInOut{Input}{Input}
    \SetKwInOut{Output}{Output}

    \Input{RGB images, corresponding pose $(\mathbf{R}, \mathbf{t})$, intrinsic $\mathbf{K}$, and depth map $\mathbf{D}$}
    \Output{3D Pseudo Boxes ${b}^{\text{3D}}$}
    \BlankLine
    \For{each RGB image $I$}{
        $n_{j}^{\text{2D}}$ ← Segment2D($I$)

        $n^{\text{3D}}_j$ ← Backproject($n_{j}^{\text{2D}}$, $(\mathbf{R}, \mathbf{t})$, $\mathbf{K}$, $\mathbf{D}$) \tcp*{\textcolor{blue}{\textnormal{Eq.\textcolor{green}{1}}}}
    }
    \BlankLine
    $Nodes$ := $\{n_{j}^{\text{3D}}\}$
    
    $V$ ← Voxelize($Nodes$) \tcp*{\textcolor{blue}{\textnormal{Voxelize based on 3D coordinates of each node}}}
    \BlankLine

    \For{voxel in $V$}{ 
        \For{any pair $(n_{j}^{\text{3D}}, n_{k}^{\text{3D}})$ in voxel}{
            $e_{jk} := \mathrm{edge}(n_{j}^{\text{3D}}, n_{k}^{\text{3D}})$ \tcp*{\textcolor{blue}{\textnormal{Eq.\textcolor{green}{2}}}}
        }
    }
    \BlankLine

    $Edges$ := $\{e_{jk}\}$

    $Embedding$ ← GraphEmbed(GenGraph($Nodes$, $Edges$))
    
    $C$ ← Clustering($Embedding$) \tcp*{\textcolor{blue}{\textnormal{Give each node a clustered group}}}

    \BlankLine

    \For{$\mathcal{C}_{q}$ in $C$}{

        $\hat{n}^{\text{3D}}_{q}$ := $\{n^{\text{3D}}\in \mathcal{C}_{q}\}$     \tcp*{\textcolor{blue}{\textnormal{Collect partial segments in the same cluster $q$}}}
        
        ${b}^{\text{3D}}_{q}$ := AxisAlignedBox($\hat{n}^{\text{3D}}_{q}$)
    }
    \BlankLine
    ${b}^{\text{3D}}$ := $\{b_{q}^{\text{3D}}\}$
    \caption{3D Pseudo Box Generation}
    \label{alg:pseudo_box_gen}
\end{algorithm}

\noindent\textbf{Voxel and 3D Volume.}
The feature volume measures 6.4 × 6.4 × 2.56 meters, with a voxel size of 0.16 meters in all three dimensions.

\subsection{3D Pseudo Box}
\noindent\textbf{3D Pseudo Box on ScanNetv2.}
\begin{table}[t]
\small
\centering
  \caption{
  \textbf{3D Pseudo Box Evaluation} on ScanNetv2. Our 3D pseudo boxes demonstrate higher quality compared to OV-3DET and SAM3D in terms of precision at IoU@0.25 and IoU@0.50. 
}
\vspace{-0.5em}
    \begin{tabular}{@{}lcccc@{}}
    \toprule
    \multirowcell{2}[-0.6ex][l]{Method} & \multicolumn{2}{c}{Precision (\%)} & \multicolumn{2}{c}{Recall (\%) } \\
    \cmidrule(l){2-3} \cmidrule(l){4-5}
    & \multicolumn{1}{c}{@0.25} & \multicolumn{1}{c}{@0.50} & \multicolumn{1}{c}{@0.25} & \multicolumn{1}{c}{@0.50}\\
    \midrule
    OV-3DET~\cite{lu2023openvocabulary} & \multicolumn{1}{c}{4.28} & \multicolumn{1}{c}{0.20} & \multicolumn{1}{c}{53.14} & \multicolumn{1}{c}{25.90} \\
    SAM3D~\cite{yang2023sam3d} & \multicolumn{1}{c}{7.39} & \multicolumn{1}{c}{4.94} & \multicolumn{1}{c}{70.02} & \multicolumn{1}{c}{\textbf{46.76}} \\
    \midrule
    Ours w/o MSR & \multicolumn{1}{c}{15.81} & \multicolumn{1}{c}{7.52} & \multicolumn{1}{c}{72.62} & \multicolumn{1}{c}{34.56} \\
    Ours & \multicolumn{1}{c}{\textbf{17.11}} & \multicolumn{1}{c}{\textbf{9.91}} & \multicolumn{1}{c}{\textbf{73.84}} & \multicolumn{1}{c}{42.80} \\
    \bottomrule
    \end{tabular}
  \label{table:scannetv2-pseudo-box}%
\end{table}%
The evaluation result for ScanNetv2~\cite{dai2017scannet} is presented in Table~\ref{table:scannetv2-pseudo-box}. Similar to the performance on ScanNet200, our method consistently outperforms OV-3DET~\cite{lu2023openvocabulary} and SAM3D~\cite{yang2023sam3d} in terms of precision at IoU@0.25 and IoU@0.50, while maintaining a comparable recall with SAM3D. 
This validates the contribution of the graph embedding-based clustering strategy, which simultaneously considers the 2D segmentation results across all frames. This approach helps mitigate the impact of segmentation errors from individual frames.

\noindent\textbf{3D Pseudo Box in Different Subset on ScanNet200.}
\begin{table}[t]
\small
\centering
  \caption{\textbf{Detailed 3D Pseudo Box Evaluation with different 2D segmentation} on ScanNet200. We perform a comprehensive evaluation across different subsets of ScanNet200. Additionally, we leverage various 2D segmentation sources to generate pseudo boxes. The use of different 2D segmentation sources in our method results in 3D pseudo boxes of varying quality. For example, when CropFormer is applied, these boxes outperform all other methods in terms of precision and recall at IoU@0.25 and IoU@0.50.}
    \resizebox{1\columnwidth}{!}{\begin{tabular}{@{}l|c|ccccc@{}}
    \toprule
    \multirowcell{2}[-0.6ex][l]{Method} & \multirowcell{2}[-0.6ex][c]{2DSeg}  & \multirowcell{2}[-0.6ex][l]{Classes} & \multicolumn{2}{c}{Precision (\%)} & \multicolumn{2}{c}{Recall (\%) } \\
    \cmidrule(l){4-5} \cmidrule(l){6-7}
    & & &\multicolumn{1}{c}{@0.25} & \multicolumn{1}{c}{@0.50} & \multicolumn{1}{c}{@0.25} & \multicolumn{1}{c}{@0.50}\\
    \midrule
    \multirowcell{4}[-0.6ex][l]{OV-3DET~\cite{lu2023openvocabulary}} & \multirowcell{4}[-0.6ex][c]{Detic~\cite{Detic}} & \multicolumn{1}{c}{overall} & \multicolumn{1}{c}{11.62} & \multicolumn{1}{c}{4.40} & \multicolumn{1}{c}{21.13} & \multicolumn{1}{c}{7.99} \\
    \cmidrule(l){3-7}
    & & \multicolumn{1}{c}{head} & \multicolumn{1}{c}{9.59} & \multicolumn{1}{c}{3.68} & \multicolumn{1}{c}{20.39} & \multicolumn{1}{c}{7.82} \\
    & & \multicolumn{1}{c}{common} & \multicolumn{1}{c}{1.95} & \multicolumn{1}{c}{0.74} & \multicolumn{1}{c}{26.12} & \multicolumn{1}{c}{9.95} \\
    & & \multicolumn{1}{c}{tail} & \multicolumn{1}{c}{1.06} & \multicolumn{1}{c}{0.29} & \multicolumn{1}{c}{24.22} & \multicolumn{1}{c}{6.77} \\
    \midrule
    \multirowcell{4}[-0.6ex][l]{SAM3D~\cite{yang2023sam3d}} & \multirowcell{4}[-0.6ex][c]{SAM~\cite{SAM}} & \multicolumn{1}{c}{overall} & \multicolumn{1}{c}{14.48} & \multicolumn{1}{c}{9.05} & \multicolumn{1}{c}{57.70} & \multicolumn{1}{c}{36.07} \\
    \cmidrule(l){3-7}
    & & \multicolumn{1}{c}{head} & \multicolumn{1}{c}{12.54} & \multicolumn{1}{c}{7.68} & \multicolumn{1}{c}{58.44} & \multicolumn{1}{c}{35.81} \\
    & & \multicolumn{1}{c}{common} & \multicolumn{1}{c}{1.86} & \multicolumn{1}{c}{1.33} & \multicolumn{1}{c}{\textbf{56.97}} & \multicolumn{1}{c}{\textbf{40.67}} \\
    & & \multicolumn{1}{c}{tail} & \multicolumn{1}{c}{0.87} & \multicolumn{1}{c}{0.59} & \multicolumn{1}{c}{\textbf{44.72}} & \multicolumn{1}{c}{\textbf{30.27}} \\
    \midrule
    \multirowcell{4}[-0.6ex][l]{Ours w/o MSR} & \multirowcell{4}[-0.6ex][c]{SAM~\cite{SAM}}  & \multicolumn{1}{c}{overall} & \multicolumn{1}{c}{27.09} & \multicolumn{1}{c}{11.98} & \multicolumn{1}{c}{52.43} & \multicolumn{1}{c}{23.18} \\
    \cmidrule(l){3-7}
    & & \multicolumn{1}{c}{head} & \multicolumn{1}{c}{24.26} & \multicolumn{1}{c}{10.67} & \multicolumn{1}{c}{54.90} & \multicolumn{1}{c}{24.14} \\
    & & \multicolumn{1}{c}{common} & \multicolumn{1}{c}{2.99} & \multicolumn{1}{c}{1.40} & \multicolumn{1}{c}{45.15} & \multicolumn{1}{c}{21.23} \\
    & & \multicolumn{1}{c}{tail} & \multicolumn{1}{c}{0.86} & \multicolumn{1}{c}{0.36} & \multicolumn{1}{c}{20.65} & \multicolumn{1}{c}{8.68} \\
    \midrule
    \multirowcell{4}[-0.6ex][l]{Ours} & \multirowcell{4}[-0.6ex][c]{SAM~\cite{SAM}}  & \multicolumn{1}{c}{overall} & \multicolumn{1}{c}{32.07} & \multicolumn{1}{c}{18.14} & \multicolumn{1}{c}{58.30} & \multicolumn{1}{c}{32.99} \\
    \cmidrule(l){3-7}
    & & \multicolumn{1}{c}{head} & \multicolumn{1}{c}{28.55} & \multicolumn{1}{c}{16.00} & \multicolumn{1}{c}{60.68} & \multicolumn{1}{c}{34.01} \\
    & &\multicolumn{1}{c}{common} & \multicolumn{1}{c}{3.66} & \multicolumn{1}{c}{2.20} & \multicolumn{1}{c}{51.88} & \multicolumn{1}{c}{31.68} \\
    & &\multicolumn{1}{c}{tail} & \multicolumn{1}{c}{1.15} & \multicolumn{1}{c}{0.69} & \multicolumn{1}{c}{26.30} & \multicolumn{1}{c}{22.68} \\
    \midrule
    \multirowcell{4}[-0.6ex][l]{Ours} & \multirowcell{4}[-0.6ex][c]{CropFormer~\cite{qilu2023high}}  & \multicolumn{1}{c}{overall} & \multicolumn{1}{c}{\textbf{35.58}} & \multicolumn{1}{c}{\textbf{22.72}} & \multicolumn{1}{c}{\textbf{62.60}} & \multicolumn{1}{c}{\textbf{39.97}} \\
    \cmidrule(l){3-7}
    & & \multicolumn{1}{c}{head} & \multicolumn{1}{c}{\textbf{31.67}} & \multicolumn{1}{c}{\textbf{19.97}} & \multicolumn{1}{c}{\textbf{65.14}} & \multicolumn{1}{c}{\textbf{41.08}} \\
    & &\multicolumn{1}{c}{common} & \multicolumn{1}{c}{\textbf{3.94}} & \multicolumn{1}{c}{\textbf{2.75}} & \multicolumn{1}{c}{55.07} & \multicolumn{1}{c}{38.53} \\
    & &\multicolumn{1}{c}{tail} & \multicolumn{1}{c}{\textbf{1.31}} & \multicolumn{1}{c}{\textbf{0.94}} & \multicolumn{1}{c}{29.77} & \multicolumn{1}{c}{21.33} \\
    \bottomrule
    \end{tabular}
}
  \label{table:sub-set-scannet200-pseudo-box}%
\end{table}%
In Table~\ref{table:sub-set-scannet200-pseudo-box}, we showcase the detailed 3D pseudo box evaluation in ScanNet200 for different subsets (head, common, tail). The evaluation computed overall precision without considering classes, given our pseudo boxes lack class information. While calculating precision in a certain subset, such as ``head,'' only ground truth boxes in head classes are considered. This may result in a lower head precision than the overall precision, as pseudo boxes overlapping with ground truth common/tail classes contribute to false positives in the head precision calculation.

Moreover, we utilize an advanced image segmentation method, CropFormer~\cite{qilu2023high}, to acquire more accurate 3D pseudo boxes. CropFormer's improved object-wise understanding reduces the risk of over-segmentation, enhancing the consistency in 2D views. This improvement benefits our 3D pseudo box generation method, resulting in less noisy 3D segments and more precise refinements.
Our method prioritizes pseudo box precision over recall for detector training, resulting in higher precision at IoU@0.25 and IoU@0.5 compared to OV-3DET and SAM3D in each subset. This superior quality is evident in our boxes generated based on both SAM and CropFormer. They also achieve significantly better recall than OV-3DET and remain comparable to SAM3D in most settings.

\subsection{Baseline Using *3R methods}
Recent 3R methods such as MV-Dust3R~\cite{tang2024mv} and VGGT~\cite{wang2025vggt} enable 3D scene reconstruction from RGB images and camera poses without requiring depth, aligning well with OpenM3D’s inference setting. To establish a baseline, we implemented a multi-stage pipeline that combines VGGT for 3D reconstruction and OVIR-3D~\cite{lu2023ovir} for open-vocabulary instance segmentation on ScanNet200. All components were executed using official implementations and default settings, with 3D boxes computed from axis-aligned segment bounds.

\begin{figure}[h]
\begin{center}
  \includegraphics[width=\linewidth]{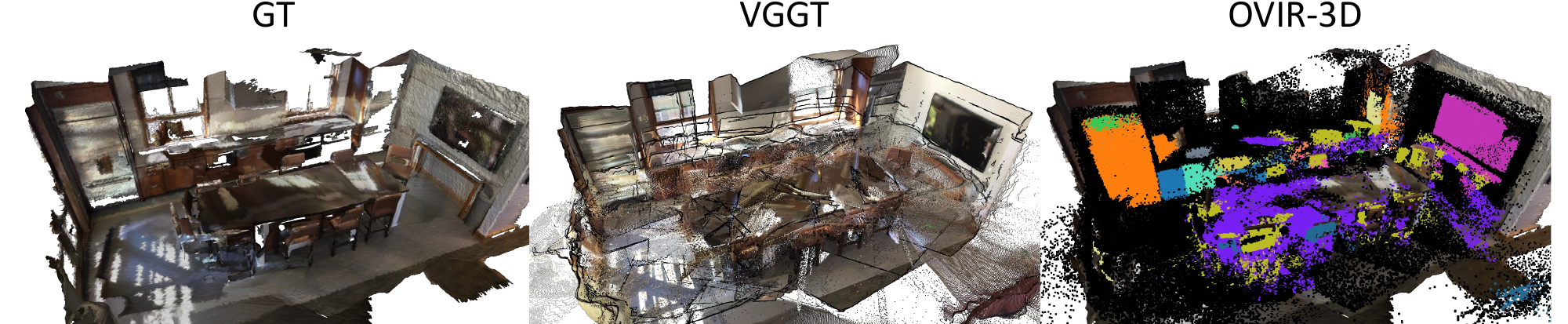}
\end{center}
\vspace{-1em}
\caption{\textbf{3R baseline qualitative result.} Comparison between (left) ground-truth ScanNet scene, (middle) VGGT 3D reconstruction using only RGB images and poses, and (right) OVIR-3D segmentation result on the VGGT output. The reconstruction lacks fine-grained indoor geometry, resulting in inaccurate 2D--3D matching and degraded segmentation quality.}
\label{fig:vggt}
\end{figure}

This pipeline incurs substantial computational overhead—particularly during 2D-3D fusion—resulting in an inference time of 300 seconds per scene, compared to 0.3 seconds for OpenM3D. In terms of accuracy, it achieved only 5.97\% AP@25 (class-agnostic), significantly lower than OpenM3D’s 26.92\%. We also observed that VGGT often fails to reconstruct fine-grained indoor geometry (see Fig.~\ref{fig:vggt}), which is crucial for accurate 2D-3D matching in instance segmentation—a limitation also noted in the OVIR-3D paper.

Overall, this reconstruction-based pipeline is substantially less effective than OpenM3D in both accuracy and efficiency for open-vocabulary 3D object detection.

\subsection{Inference Efficiency}
As shown in Table~\ref{tab:inference_time_comparison}. OpenM3D achieves the fastest inference time of 0.3 seconds per scene, using only multi-view RGB images, and significantly outperforms baselines such as OV-3DET (5 s), S2D (2.1 s), and S2D with depth estimation (81 s). Unlike others, it avoids costly CLIP inference and depth prediction, making it highly suitable for real-time 3D detection.
\begin{table}[htbp]
\centering
\caption{\textbf{Inference time comparison} on ScanNet200 on a V100 GPU. OpenM3D is over 16× faster than OV-3DET and 270× faster than the depth-estimated S2D baseline.}
\vspace{-1em}
\label{tab:inference_time_comparison}
\renewcommand{\arraystretch}{1.2}
\resizebox{\columnwidth}{!}{%
\begin{tabular}{lcccc}
\hline
\textbf{Method} & \textbf{OV-3DET} & \textbf{S2D} & \textbf{S2D Depth Est.} & \textbf{Ours} \\
\hline
Inference time (s) & 5 & 2.1 & 81 & \textbf{0.3} \\
\hline
\end{tabular}%
}
\end{table}

\subsection{Transferability of Pretrained Model}
OpenM3D does not rely on predefined `seen' categories or 3D annotations during training, making it naturally OV - all categories are essentially novel. OpenM3D demonstrates strong performance across head, common, and tail classes in ScanNet200 (see Fig.~\ref{fig:novel_tail}), highlighting its ability to handle rare or unseen classes.

\subsection{Ablation Study}
\noindent\textbf{CLIP Visual Encoders.}
We aligned our voxel feature to the pre-trained CLIP feature extracted by the ViT-L/14 image encoder during training.
Furthermore, we showcase alternative results employing various other CLIP image encoders in this section. As outlined in Table~\ref{table:scannet200-result-vit}, the use of different CLIP image encoders exhibited negligible impact on both evaluation metrics, namely mAP@25 and mAR@25. This observation emphasizes the robust open-vocabulary classification capability of our method, {\ourmethod}.
\begin{table}[t]
\small
\centering
\vspace{-1em}
\caption{
  \textbf{Results of {\ourmethod} trained with different CLIP encoders} on ScanNet200.
}
\begin{tabular}{l|cc}
    \toprule
    CLIP Encoder & mAP@25 (\%) & mAR@25 (\%) \\
    \midrule
    ViT-L/14 & 4.23 & 15.12\\
    ViT-B/16 & 4.16 & 15.50\\
    ViT-B/32 & 4.02 & 14.74 \\
    \bottomrule
\end{tabular}
\label{table:scannet200-result-vit}
\end{table}

\noindent\textbf{3D Detection with Pseudo Box using CropFormer.}
When deploying better segmentation models, e.g., CropFormer~\cite{qilu2023high}, we can generate more accurate pseudo boxes as detailed in Table~\ref{table:sub-set-scannet200-pseudo-box}. The improvement on 2D segmentation benefits our 3D pseudo box generation method on 3D segments refinements. Trained with these boxes, {\ourmethod} demonstrates a notable improvement of 12.5\% in mAP@25, rising from 4.23\% to 4.76\% on ScanNet200, as shown in Table~\ref{table:scannet200-cropformer-result}. \ke{This highlights the potential of our 3D pseudo boxes on better 2D segmentation.}
Note that in ARKitScenes, given the sparse point cloud, improving 2D segmentation using CropFormer alone has not significantly improved 3D box metric performance.

\noindent\textbf{3D Detection on ScanNetv2.}
We reported the results of our model evaluated on the common 18 classes in ScanNetv2~\cite{dai2017scannet} in Table~\ref{table:scannet-result}. {\ourmethod} trained with our pseudo boxes consistently outperforms the models trained with SAM3D and OV-3DET on all metrics, including AP@25, AP@50, AR@25, and AR@50. Notably, {\ourmethod} achieved over 12\% and 20\% improvements in AP@25 and AR@50, respectively, compared to OV-3DET. Larger gaps were observed, with 7.34\% vs. 2.87\% in AP@50 and 20.94\% vs. 9.27\% in AR@50. The substantial improvements brought by our method on AP@50 and AR@50 underscore the limitations associated with solely relying on single-view depth maps and images for bounding box generation. The notable improvements of our pseudo boxes over SAM3D in AP@25 and AP@50 metrics showcase the efficacy of our graph embedding-based pseudo boxes. Note that the train/evaluate split applied in \cite{lu2023openvocabulary, cao2023coda} differs from the official split by ScanNetv2~\cite{dai2017scannet}, making direct comparisons with their reported results challenging.

\begin{table}[t]
    \centering
    \small
    \caption{\textbf{3D Object Detection} on ScanNet200. Our pseudo boxes with CropFormer improve upon SAM.}
    \vspace{-1em}
    \begin{tabular}{rccc}
    \toprule
    \multicolumn{2}{c}{Trained Box} & \multirow{2}{*}{mAP@25(\%)} & \multirow{2}{*}{mAR@25(\%)} \\
\cline{1-2}    \multicolumn{1}{l}{Method} & 2DSeg &       &  \\
    \hline
    \multicolumn{1}{l}{OV-3DET~\cite{lu2023openvocabulary}} & Detic~\cite{Detic} & 3.13  & 10.83 \\
    \multicolumn{1}{l}{SAM3D~\cite{yang2023sam3d}} & SAM~\cite{SAM}   & 3.92  & 13.33 \\
    \hline
    \multicolumn{1}{l}{\multirow{2}[2]{*}{Ours}} & SAM~\cite{SAM}   & 4.23  & \textbf{15.12} \\
          & CropFormer~\cite{qilu2023high} & \textbf{4.76} & 14.62 \\
    \bottomrule
    \end{tabular}%
    \label{table:scannet200-cropformer-result}
\end{table}

\begin{table}[t]
    \centering
    \small
    \caption{\textbf{3D Object Detection} on ScanNetv2. {\ourmethod} outperforms our method trained on the boxes from OV-3DET and SAM3D.}
\vspace{-1em}
\resizebox{1\columnwidth}{!}{\begin{tabular}{l|c|cccc}
    \toprule
    Method & Trained Box & AP@25 (\%) & AR@25 (\%) & AP@50 (\%) & AR@50 (\%)\\
    \midrule
    \multirow{3}{*}{\ourmethod} \
     & OV-3DET~\cite{lu2023openvocabulary} & 17.65 & 40.37 & 2.87 & 9.27\\
     & SAM3D~\cite{yang2023sam3d} & 16.69 & 49.39 & 5.18 & 19.33 \\
     & Ours & \textbf{19.76} & \textbf{50.40} & \textbf{7.34} & \textbf{20.94}\\
    \bottomrule
\end{tabular}%
}
\label{table:scannet-result}
\end{table}

\section{Limitation}
The gap between class-agnostic and OV 3D detection implies that the pre-trained CLIP feature can be improved in classifying many semantically similar household objects. We leave this as a future direction.

\end{document}